\def\paperID{15317}
\def\confName{CVPR}
\def\confYear{2024}

\def\paperTitle{Self-Training Large Language Models for Improved Visual Program Synthesis With Visual Reinforcement}

\def\authorBlock{
Zaid Khan$^{1}$ \quad Vijay Kumar BG$^{2}$ \quad Samuel Schulter$^2$ \quad Yun Fu$^1$  \quad Manmohan Chandraker$^{2,3}$ \\
\small{$^1$Northeastern University \quad $^2$NEC Laboratories America \quad $^3$UC San Diego} \\
}

\newif\ifreview 
\newif\ifarxiv \newcommand{\arxiv}{\arxivtrue}
\newif\ifcamera 
\newif\ifrebuttal 

\arxiv
\pdfoutput=1
\documentclass[10pt,twocolumn,letterpaper]{article}
\usepackage[pagenumbers]{cvpr}

\usepackage{graphicx}	
\usepackage{amsmath}	
\usepackage{amssymb}	
\usepackage{booktabs}
\usepackage{times}
\usepackage{microtype}
\usepackage{epsfig}
\usepackage[table,xcdraw,dvipsnames]{xcolor}
\usepackage{caption}
\usepackage{float}
\usepackage{placeins}
\usepackage{color, colortbl}
\usepackage{stfloats}
\usepackage{enumitem}
\usepackage{tabularx}
\usepackage{xstring}
\usepackage{multirow}
\usepackage{xspace}
\usepackage{url}
\usepackage{subcaption}
\usepackage{xcolor}
\usepackage[hang,flushmargin]{footmisc}
\usepackage{listings}
\usepackage{xcolor}
\usepackage{multicol}

\ifcamera \usepackage[accsupp]{axessibility} \fi

% ...

\ifarxiv  \fi

\newcommand{\R}[1]{{%
    \textbf{%
        \ifstrequal{#1}{1}{\textcolor{red}{R#1}}{%
        \ifstrequal{#1}{2}{\textcolor{blue}{R#1}}{%
        \ifstrequal{#1}{3}{\textcolor{magenta}{R#1}}{%
        \ifstrequal{#1}{4}{\textcolor{teal}{R#1}}{%
                           \textcolor{cyan}{R#1}%
        }}}}%
    }%
}}
\usepackage{xr-hyper}

\makeatletter
\newcommand*{\addFileDependency}[1]{
  \typeout{(#1)}
  \@addtofilelist{#1}
  \IfFileExists{#1}{}{\typeout{No file #1.}}
}

\makeatother

\definecolor{cvprblue}{rgb}{0.21,0.49,0.74}
\usepackage[pagebackref,breaklinks,colorlinks,citecolor=cvprblue]{hyperref}
\usepackage[capitalize]{cleveref}
\crefname{section}{Sec.}{Secs.}
\crefname{table}{Table}{Tables}
\crefname{figure}{Fig.}{Figs.}

\usepackage{colortbl}
\usepackage{booktabs}
\usepackage{tikz}
\usepackage{adjustbox}
\definecolor{lightgray}{gray}{0.9}
\definecolor{yesColor}{rgb}{0.0, 0.5, 0.0}
\definecolor{noColor}{rgb}{0.8, 0.0, 0.0}
\definecolor{strongColor}{rgb}{0.8, 0.0, 0.0}
\definecolor{weakColor}{rgb}{0.0, 0.5, 0.0}

\newcommand{\yes}{\textcolor{yesColor}{Yes}}
\newcommand{\no}{\textcolor{noColor}{No}}
\newcommand{\strong}{\textcolor{strongColor}{Strong}}
\newcommand{\weak}{\textcolor{weakColor}{Weak}}
\newcommand{\ourmethod}{\textbf{VisReP}}

\begin{document}
%% TITLE
\title{\paperTitle}
\author{\authorBlock}
\maketitle

\begin{abstract}
Visual program synthesis is a promising approach to exploit the reasoning abilities of large language models for compositional computer vision tasks. 
Previous work has used few-shot prompting with frozen LLMs to synthesize visual programs.
Training an LLM to write better visual programs is an attractive prospect, but it is unclear how to accomplish this.
No dataset of visual programs for training exists, and acquisition of a visual program dataset cannot be easily crowdsourced due to the need for expert annotators.
To get around the lack of direct supervision, we explore improving the program synthesis abilities of an LLM using feedback from interactive experience.
We propose a method where we exploit existing annotations for a vision-language task to improvise a coarse reward signal for that task, treat the LLM as a policy, and apply reinforced self-training to improve the visual program synthesis ability of the LLM for that task. We describe a series of experiments on object detection, compositional visual question answering, and image-text retrieval, and show that in each case, the self-trained LLM outperforms or performs on par with few-shot frozen LLMs that are an order of magnitude larger. Website: \url{https://zaidkhan.me/ViReP}
\end{abstract}
\section{Introduction}
\label{sec:intro}
\begin{figure}[t]
    \centering
    \includegraphics[width=\linewidth]{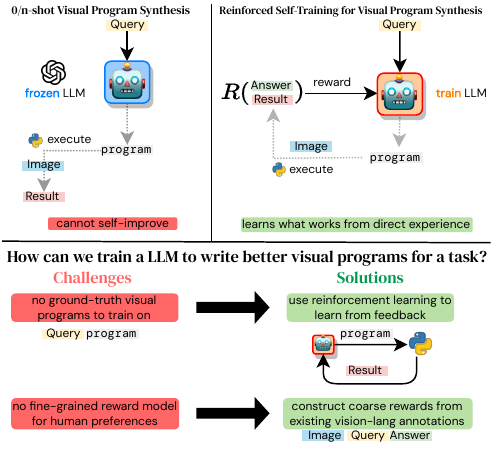}
    \caption{Visual program synthesis with LLMs has been treated as a 0/n-shot task where the LLM is kept frozen. This limits opportunities for improvement. We ask whether it is possible to train a LLM to write more accurate programs. Given that there is no large scale dataset of accurate visual programs available, we propose improving the LLM using self-training.}
    \label{fig:teaser}
\end{figure}
\begin{table*}[t]
\centering
\resizebox{\linewidth}{!}{
\begin{tabular}{ccccccccc}
\toprule
 & \textbf{Task Domain} & \textbf{Self-Training} & \textbf{Supervision} & \textbf{Tool/API Use} & \textbf{Visual Task Decomposition} & \textbf{Grounded By Feedback} & \textbf{Improves LLM} \\
\midrule
\ourmethod~(Ours) & \textcolor{blue}{Visual Program Synthesis} & \yes & \weak & \yes & \yes & \yes & \yes \\
\citet{LangModelsCanKalai2023} & \textcolor{brown}{Programming Puzzles} & \yes & \weak & \no & \no & \yes & \yes \\
ReST \cite{rest} & \textcolor{gray}{Natural Language Understanding} & \yes & - & \no & \no & \yes & \yes \\
VisProg \cite{visprog} & \textcolor{blue}{Visual Program Synthesis} & \no & - & \yes & \yes & \no & \no \\
ViperGPT \cite{vipergpt} & \textcolor{blue}{Visual Program Synthesis} & \no & \weak & \yes & \yes & \no & \no \\
ToolLLM \cite{toolllm} & Tool Usage by API & \no & \strong & \yes & \no & \no & \yes \\
GorillaLLM \cite{gorillallm} & Tool Usage by API & \no & \strong & \yes & \no & \no & \textcolor{yesColor}{Yes} \\
ToolFormer \cite{toolformer} & \textcolor{gray}{Natural Language Understanding} & \yes & \weak & \yes & \no & \yes & \textcolor{yesColor}{Yes} \\
\bottomrule
\end{tabular}

}
\caption{
Differences between our work and similar work.
Strong supervision means that the training process requires examples of ground-truth programs to train the LLM.
Weak supervision means that the training process does not require ground-truth programs.
Tool / API use means that the LLM is required to use substantial functionality implemented by external modules (e.g. an object detector, a web search API) to solve tasks.
Visual task decomposition means that the LLM can decompose a complex visual task into primitive subtasks.
Grounded by feedback means that the LLM has been optimized not just for syntactic / semantic correctness (program does not hallucinate / cause errors), but for functional correctness (programs produce the correct answer).
Improves LLM means that the work proposes a method to improve an LLM for a specific task, rather than using a frozen LLM.
}
\vspace{-0.05in}
\label{tab:comparison}
\end{table*}

Complex visual queries can often be decomposed into simpler subtasks, many of which can be carried out by task-specific \textit{perception modules} (e.g. object detection, captioning). 
For example, consider the problem of finding bounding boxes for the phrase ``white mug to the left of the sink''.
This is a challenging query for single model such as an open vocabulary object detector.
However, this query can be solved by writing a program that composes task-specific perception modules with logic: use an open vocabulary object detector to find a sink and white mugs in the scene, then compare the horizontal center of the sink and the mugs to find white mugs to the left of the sink.
% However, a human could write a formal program that invokes task-specific modules interleaved with logic.
Program synthesis with large language models \cite{program_synthesis_with_llms} is a promising approach to automate this process, and recent work has shown that proprietary large language models can write programs for visual tasks \cite{vipergpt, visprog, modular_vqa_via_cg}.
Current approaches for visual program synthesis with LLMs use few-shot prompting and rely on the in-context learning abilities \cite{icl_bayesian} of frozen, proprietary LLMs. (\cref{fig:teaser})

Few-shot prompting with frozen LLMs for visual program synthesis as in ViperGPT \cite{vipergpt}, VisProg \cite{visprog}, or CodeVQA \cite{modular_vqa_via_cg} has several limitations.
The LLM needs to understand the competencies of the perception modules it is using.
A open vocabulary object detector may able to locate a common attribute-noun phrase such as ``white mug'' without problems, but struggle with a more abstract phrase such as ``microwaveable mug'' \cite{omnilabel}.
A VQA model might be able to answer ``is the car blue?'' without problems, but fail when logical modifiers are introduced, such as ``is the car not blue?'' \cite{vqa_lol}.
In many cases, we do not precisely know the weaknesses or competencies of a perception model \cite{reliable_vqa}. 
Even if this were known, it is difficult to convey \textit{all} of the competencies / weaknesses of a perception module through in-context examples.
Second, it is the case in program synthesis that an LLM often \textit{can} generate the correct solution to a problem, but the correct solution is not the solution the LLM places the highest probability on \cite{chen2021codex}.
We would like to align the LLM to ``uncover'' the knowledge of the correct solution, but it not clear how to do this in a principled way with few-shot prompting alone.
\begin{quote}
    \textit{How can we train a large language model to write better visual programs for a specific task?}
\end{quote}

\textbf{Our goal is to optimize the parameters of the language model so the accuracy of the synthesized programs is higher.}
Existing approaches that train LLMs to improve their ability to programatically use tools / APIs such as GorillaLLM \cite{gorillallm}, ToolLLM \cite{toolllm} do so by finetuning LLMs on examples of tool use or API use.
This cannot be directly applied to visual program synthesis because \textbf{there are no large scale datasets of visual programs, and collecting such a dataset would be extremely labor intensive.}
In the absence of a large scale dataset, how do we learn to write better programs for a visual task?
\begin{quote}
    \textit{We posit that grounding a language model with interactive feedback from a generic visual task will improve the general visual program synthesis abilities of the model.}
\end{quote}
% ToolFormer \cite{toolformer} learns to use tools in a self-supervised manner, but their method is specific to natural language understanding tasks and does not synthesize programs.
A natural way to learn from feedback is to use reinforcement learning. 
ReST \cite{rest} and RaFT \cite{raft} introduce a general framework for reinforced self-training in generative tasks and demonstrate success in machine translation and text-to-image generation.
However, a crucial ingredient in their recipe is the availability of a fine-grained reward model.
It is difficult to construct a fine-grained reward model for visual program synthesis, given both the absence of human preference datasets for visual programs, and the difficulty of devising a proxy metric.
One alternative is to use unit tests to teach a neural reward model or give a coarse-grained reward.
This technique has been used successfully in coding challenges by CodeRL \cite{coderl} and \citet{LangModelsCanKalai2023}, but it is unclear how it can be applied to visual program synthesis.
Our key idea is to use existing annotations for a vision-language as improvised unit tests to provide a coarse reward signal. 
Using the coarse reward signal, we can apply reinforced self-training by treating the language model as a policy and training it with a simple policy gradient algorithm.
We alternate synthetic data generation steps in which we sample programs from the language model policy with optimization steps in which we improve the language model policy based on observations from executing the sampled programs.
% We conduct experiments on compositional visual question answering, object detection, and image-text matching.
We name our proposed method \ourmethod, for Visually Reinforced Program Synthesis.

\begin{itemize}
    \item We propose \textit{optimizing} the parameters of a LLM so that the accuracy of the synthesized visual programs is higher, in contrast to previous works that use frozen LLMs.
    \item Since no dataset of accurate visual programs is available for finetuning, we hypothesize that we can instead use feedback from the execution environment to improve the visual program synthesis abilities of a language model.
    % \item We show that relatively tiny amount of human annotations improves the stability of the self-training process and capabilities of the language model.
    \item We propose \ourmethod, an offline, model agnostic recipe for reinforced self-training of large language models for visual program synthesis using existing vision-language annotations with a simple policy gradient algorithm.
    \item Our results show that it is possible to apply reinforced self-training for to improve large language models for visual program synthesis \textit{with} only coarse rewards. 
\end{itemize}

We demonstrate the effectiveness of an CodeLlama-7B policy trained by \ourmethod~on compositional visual question answering ($+9\%$), complex object detection ($+5\%$), and compositional image-text matching ($+15\%$) relative to the untrained policy.
We show that the policy trained by \ourmethod~exceeds the accuracy of a gpt-3.5-turbo policy on all three tasks.
\section{Related Work}
\label{sec:related}
\subsection{Self-Training}
Self-training is an established paradigm which uses unlabeled data to improve performance.
Self-training has been successfully applied in a number of fields.
We restrict our coverage to usages with significant overlap.

\noindent \textbf{Program Synthesis} \citet{LangModelsCanKalai2023} showed that LLMs can improve their program synthesis abilities by generating programming puzzles and solving them.
CodeRL \cite{coderl} proposed an actor-critic framework to improve the program synthesis abilities of LLMs for programming problems accompanied by unit tests.
CodeIT \cite{codeit} and Rest-EM \cite{rest_em} also use a similar policy gradient approach for program synthesis.
Our problem domain is different from these works, which focus on program synthesis for programming puzzles / problems.
In addition, our work has an explicit focus on learning to use an API fluently.

\noindent \textbf{Alignment} ReST \cite{rest} and RaFT \cite{raft} introduced a generic framework for reinforced self-training and applied it to align machine translation outputs to human preferences and align foundation models on language understanding and image generation tasks respectively.
These works share the same basic idea as our work, though they are in a substantially different task domain where human preferences are either known (conversational alignment) or can be estimated with an available neural model.

\noindent \textbf{Vision-Language} SelTDA \cite{seltda} introduced a self-training approach for visual question answering.
SelTDA proceeds by pseudolabeling unlabeled data, then finetuning a large VLM on the pseudolabeled data.
In contrast to SelTDA, we improve a LLM for visual program synthesis.

\subsection{Visual Program Synthesis}
Visual program synthesis with LLMs was proposed concurrently by ViperGPT \cite{vipergpt}, VisProg \cite{visprog}, and CodeVQA \cite{modular_vqa_via_cg}.
The common points between these three works is that (a) they use pretrained LLMs as code generators (b) they represent complex visual tasks as compositions of primitive visual subtasks (c) they use code to invoke task-specific models to perform the primitive subtasks.
Our work is most similar to ViperGPT and CodeVQA as they produce code in a general purpose programming language rather than a DSL.
All three works use a proprietary, frozen LLM.
In contrast to all three, the focus of our work is on how we can improve the visual program synthesis abilities of an open LLM.

\subsection{Tool Use with LLMs}
Multimodal tool-using LLMs were first introduced by Socratic Models \cite{socratic_models}.
However, their approach was to create fixed pipelines in which the output of a perception model such as CLIP \cite{clip} is fed to a LLM. 
Later approaches such as GorillaLLM \cite{gorillallm} and ToolLLM \cite{toolllm} improved on this by treating tool use as a program synthesis problem and creating LLMs that use a broad range of tools by learning to invoke APIs.
However, one key limitation of these approaches in the context of visual program synthesis is that that they do not learn to \textit{decompose} problems into subproblems that can be solved by tools.
Instead, they are trained to select the right tool for the problem and invoke it.
Another limitation is that they are not optimized for functional correctness.
They are trained for syntactic and semantic correctness, but they have not been provided feedback on whether their use of tools produces the desired answer.
ToolFormer \cite{toolformer} is similar to our work in the sense that the LLM's usage of tools is grounded by feedback, but they focus on natural language understanding tasks rather than visual tasks.

\section{Method}
\subsection{Visual Program Synthesis with LLMs}
\textbf{Task Formulation} Let $v$ be a visual input and $q$ be a textual query about $v$.
In visual program synthesis, we synthesize a program $p=\pi_\theta (q)$ with a program generator $\pi_\theta$.
The program $p$ and visual input $q$ are then fed into the execution engine $\hat{y}=\phi(v,p)$ to produce a result $\hat{y}$.
The program generator is an auto-regressive large language model
\begin{equation}
\pi_\theta(\boldsymbol{y} \mid \boldsymbol{x})=\prod_{t=1}^T \pi_\theta\left(p_t \mid \mathbf{p}_{1: t-1}, \boldsymbol{x}\right),
\end{equation}
where $\mathbf{p}_{1:t}$ are the tokens of the program, and $x$ is the input to the large language model.
The language model is kept frozen in previous work \cite{vipergpt}.
Our goal is to optimize the parameters $\theta$ of the language model $\pi$ so the accuracy of the synthesized programs is higher.

\noindent\textbf{Implementation} Following ViperGPT \cite{vipergpt}, we provide the specification of the \texttt{ImagePatch} API concatenated with the textual query $q$ as the prompt to the program generator. 
The synthesized program $p$ is a Python program that can invoke any Python builtins, control flow structures, and the \texttt{ImagePatch} API.
Our implementation of the \texttt{ImagePatch} API is largely similar to ViperGPT.
We remove some API methods that were not required for the tasks we evaluate on (such as \texttt{llm\_query}).
We use BLIP \cite{blip} and GroundingDINO \cite{groundingdino} as perception modules underlying \texttt{find} (object detection), \texttt{simple\_query} (visual question answering), and \texttt{verify\_property} (attribute verification).
\subsection{Reinforced Self-Training}
\label{sec:rest}
\begin{figure}[ht]
    \centering
    \includegraphics[width=\linewidth]{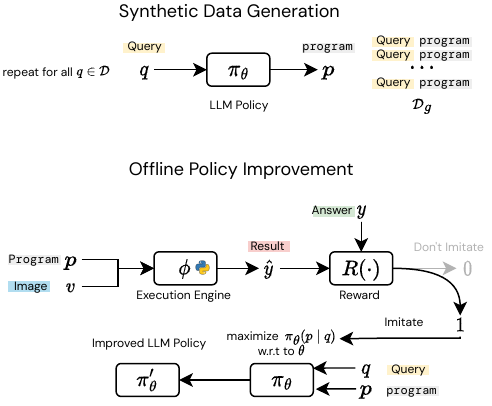}
    \caption{\ourmethod~can be applied to improve the visual synthesis abilities of an LLM for a vision-language task using existing annotations for a vision-language task (e.g. an object description+image+bounding boxes). A key idea is to construct a coarse reward by comparing the answer produced by a synthesized program to the ground-truth answer.}
    \label{fig:architecture}
\end{figure}
Rather than use a frozen large language model as the program generator $p_\theta$, we would like to optimize the parameters  $\theta$ of the language model so the accuracy of the synthesized programs is higher.
It is not obvious how to do this.
We can't backpropagate through the execution engine $\phi(\pi_\theta(q),v)$ to directly optimize $\theta$ with respect to $q$ or $v$.
An alternative might be to use human labor to build a dataset of high-quality visual programs, and train the large language model $\pi_\theta$ on the manually-collected dataset.
But collecting such a dataset is very labor intensive, and not scalable.
Instead, we explore the idea of learning from experience by applying a simple policy gradient method, REINFORCE \cite{reinforce}.

We propose \ourmethod, which treats the program synthesis task as a growing batch RL problem \cite{growing_batch_rl}, inspired by ReST \cite{rest}.
We first define a coarse discrete reward function $R(\cdot)$ from existing annotations for a vision-language task.
We then alternate \textbf{Grow} steps, in which we sample trajectories (programs) from the policy (large language model), with \textbf{Improve} steps, in which we apply behavioral cloning with a reward-weighted negative-log likelihood loss to improve the policy.
A diagram of our approach is depicted in \cref{fig:architecture}.

\textbf{Grow Step} The grow step corresponds to the acting step in reinforcement learning, and can also be seen as synthetic data generation.
Let $\mathcal{D} = \{(v_1, q_1, y_1), \ldots (v_n, q_n, y_n)\}$ be a dataset for a vision-language task, where $v_i$ is an image, $q_i$ is a textual query, and $y_i$ is ground-truth for the $i$-th triplet (e.g. a string for VQA, bounding boxes for object detection).
We start with the frozen language model $\pi_\theta(p \mid q)$, where $p$ is a synthesized program and $q$ is a textual query.
The language model $\pi_\theta$ represents our policy.
We generate a dataset of trajectories $\mathcal{D}_g$ by sampling many programs $p$ from the current policy $\pi_\theta$:  $p \sim \pi_\theta(p \mid q)$ for $q \sim \mathcal{D}$. 

\textbf{Improve Step} Our goal in this step is to use the dataset of synthetic programs $\mathcal{D}_g$ to improve the policy $\pi_\theta$.
First, we define a binary-valued reward function $R: p,v,y \rightarrow \{0,1\}$ on a given program, image, annotation triplet,
\begin{equation}
    R(v, p, y) = \begin{cases} 
        1, & \text{if } \phi(p,v) = y \\
        0, & \text{otherwise}
\end{cases}
\end{equation}
where $\phi(p,v)$ is the result of executing the program $p$ on an image $v$.
Note that $y$ is \textbf{not} a program but an existing annotation such as a string for VQA for a bounding box for object detection.
To apply behavioral cloning, we then minimize the reward-weighted loss
\begin{equation}
J(\theta)=\mathbb{E}_{(q,p) \sim \mathcal{D}_g}[R(v, p) \mathcal{L}(p, q ; \theta)]
\end{equation}
where $\mathcal{L}(p,q; \theta)$ is the negative log-likelihood loss
\begin{equation}
\mathcal{L}_{\mathrm{NLL}}(p,q;\theta)=-\mathbb{E}_{(q, p) \sim \mathcal{D}_g}\left[\sum_{t=1}^T \log \pi_\theta\left(p_t \mid p_{1: t-1}, q\right)\right]
\end{equation}
over the pairs of textual queries $q$ and synthetic programs $p$  in $\mathcal{D}_g$.

Because the reward function only takes on binary values, we can simplify this and implement it by:
First, generating a dataset of synthetic programs $\mathcal{D}_g = \{\pi_\theta(q): \forall q \in \mathcal{D}\}$ using the LLM $\pi_\theta$ on a dataset $\mathcal{D}$.
Next, filtering $\mathcal{D}_g$ to obtain $\mathcal{D}_g^\prime = \{(q,v,p \in \mathcal{D}_g: R(q,v,p) > 0  \}$, which corresponds to executing all synthetic programs and only keeping those that give correct answers.
Finally, we finetune the language model $\pi_\theta$ on the filtered dataset $\mathcal{D}_g^\prime$ using the standard causal language modeling loss.
We then iterate the process, initiating a new synthetic data generation step with the improved policy $\pi_\theta^\prime$.

\textbf{Iteration} For the initial grow step, we use a frozen language model as the initial policy. 
For example, we use the pretrained \texttt{codellama-7b-instruct-hf} as the policy in the initial grow step.
In subsequent steps, we use the policy trained in the previous improve step for the grow step.

\section{Understanding Self-Training}
\begin{figure*}
    \centering
    \includegraphics[width=\textwidth]{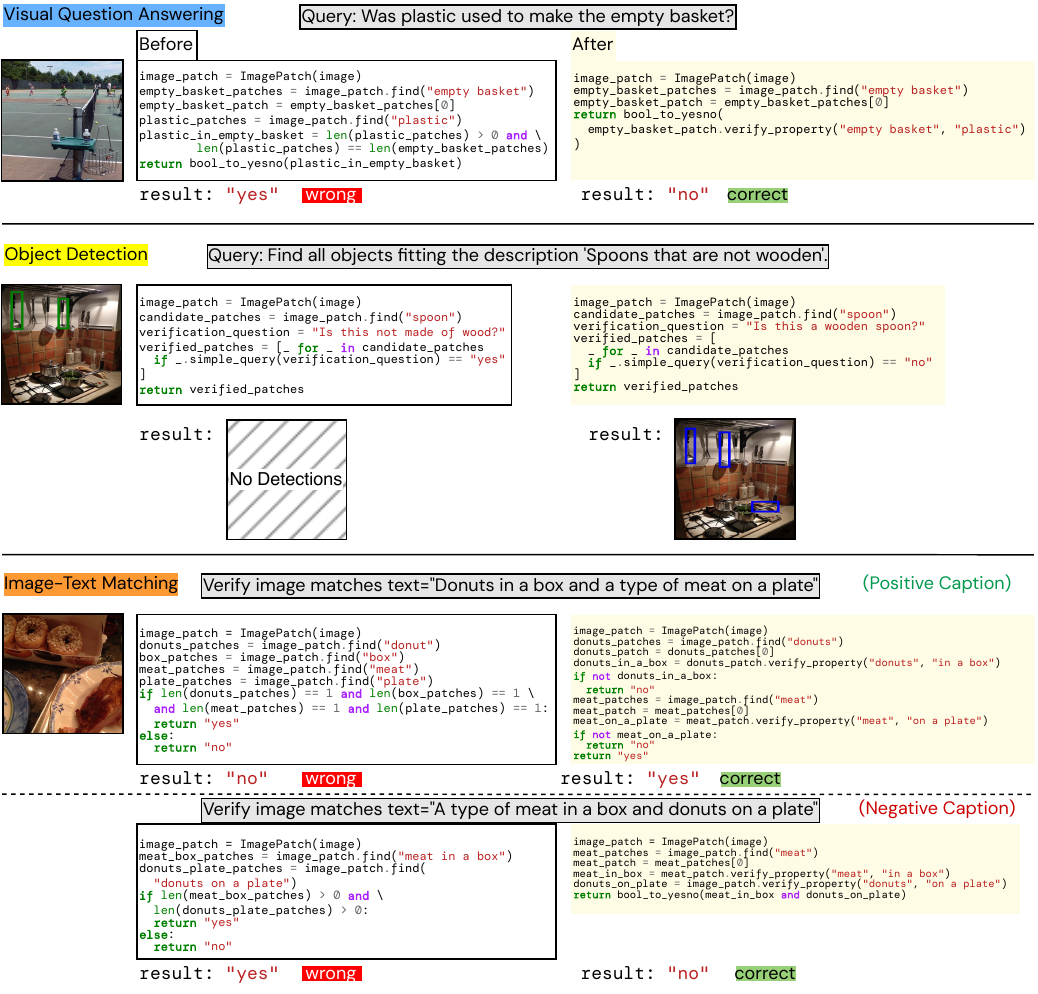}
    \caption{Self-training with \ourmethod~produces qualitatively better programs. Here, we show programs written by the initial policy (on the left) and the policy after 10 iterations of self-training on GQA (on the right). In VQA example, the initial policy does not specifically check whether the empty basket is plastic. In the object detection example, the reasoning of the initial policy is correct, but it issues a confusingly worded query to the \texttt{simple\_query} module, which returns the wrong answer. The learned policy uses \texttt{simple\_query} more appropriately. In the image-text matching example, in the initial policy tries to use the object detector to search directly for ``meat in a box'' and ``donuts on a plate'', but this is too complicated for the object detector to localize. After self-training, the LLM policy no longer makes this mistake.}
    \label{fig:before-after}
\end{figure*}
Our goal in this section is to characterize the stability and sample efficiency of \ourmethod.
We want to understand:
\begin{enumerate}
    \item How does applying \ourmethod~change the accuracy of synthesized programs?
    \item What happens as \ourmethod~is repeated?
    \item How does data scarcity and diversity affect \ourmethod?
\end{enumerate}

\subsection{Implementation}
\label{sec:impl}
We start off with the GQA \cite{gqa} dataset for visual question answering.
We choose GQA because each question in GQA was constructed programatically and is thus a good candidate to be answered by program synthesis.
GQA has over 2M questions, each belonging to one of $\approx 100$ question types.
We construct a training set by sampling $100$ questions for each question type, for a total of $\approx 10$k visual questions and answers.
We construct a validation set following \citet{visprog}.
We use the CodeLlama \cite{codellama} family of models as our initial policy.
We use LoRA \cite{lora} adapters during the \textbf{Improve} steps.
We use the hyperparameters suggested by \citet{qlora}.
Full implementation details are in the supplement.
\begin{figure*}[ht]
\centering
\includegraphics[width=\textwidth]{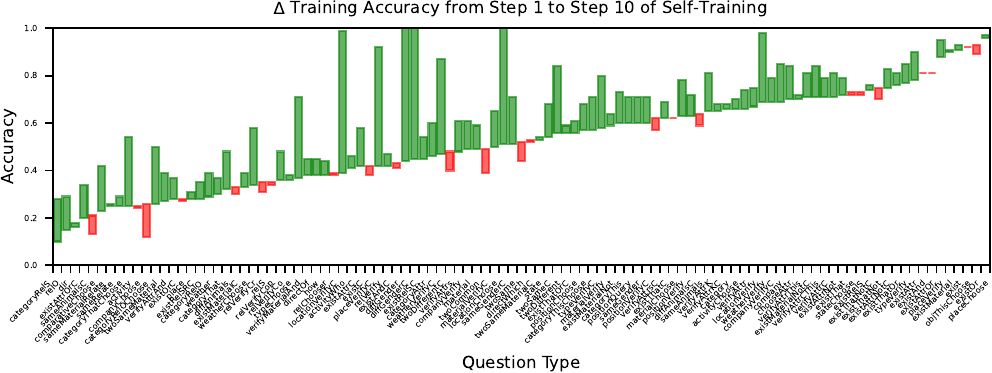}
\caption{Iteratively applying \ourmethod~allows a LLM to self-improve improve on almost all of GQA's $\approx$ 100 question types. The base of each bar is set to the accuracy of the initial policy (codellama-7b-instruct). A \textcolor{ForestGreen}{green bar} indicates question types on which the policy at iteration 10 improved over the initial policy, and a \textcolor{red}{red bar} indicates question types on which the policy at iteration 10 was worse than the initial policy.}
\label{fig:question-types-before-after-self-training}
\end{figure*}

\begin{figure}[ht]
    \centering
    \includegraphics[width=\linewidth]{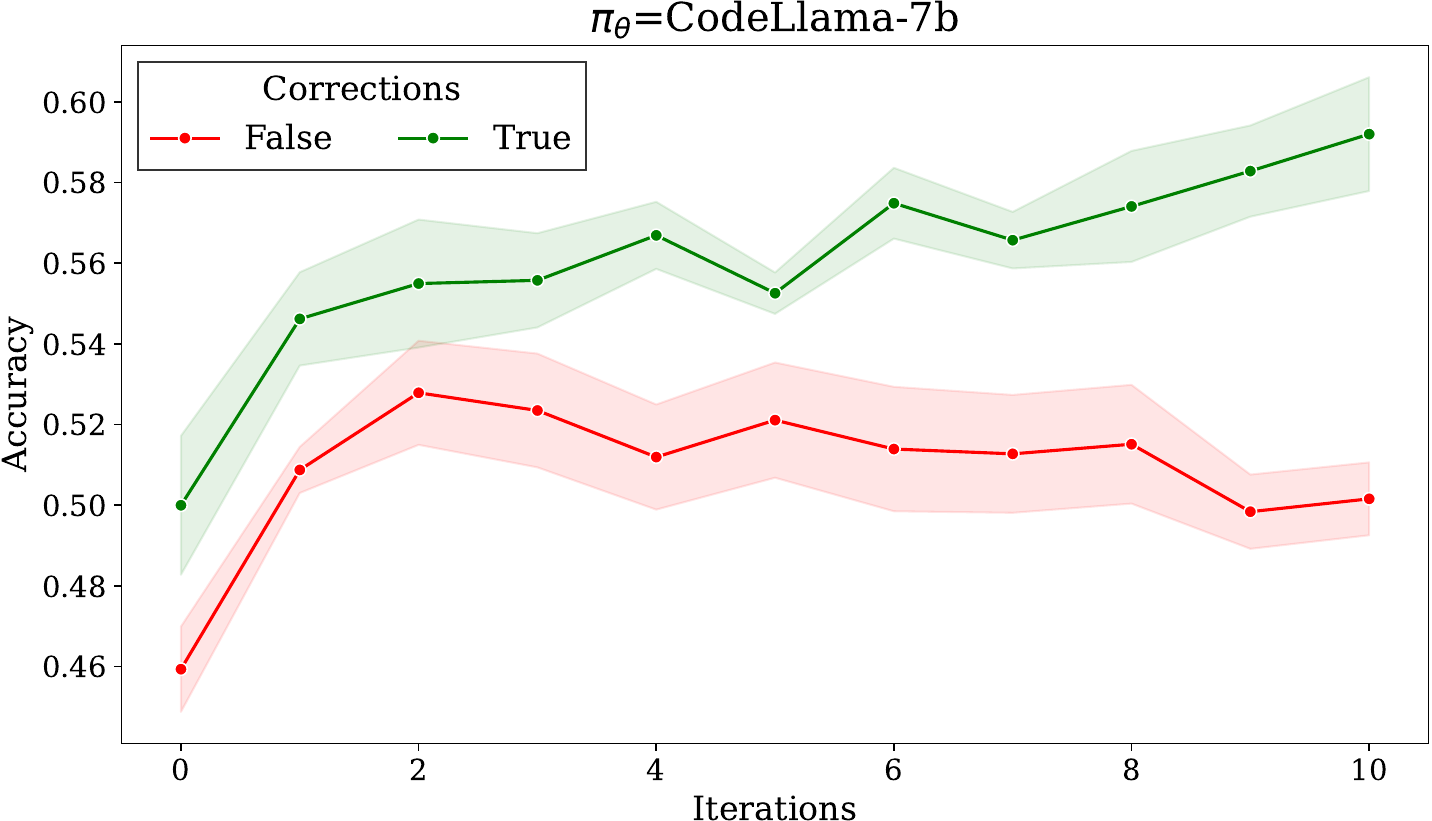}
    \caption{Supplying a small amount of human written corrections as in-context examples during training can increase the stability of the self-training process (\textcolor{ForestGreen}{green line}). We show validation accuracy on GQA through multiple iterations of self-training with a policy instantiated from CodeLlama-7b. Without these corrections, proliferating errors cause performance to degrade in later iterations (\textcolor{red}{red line}). The translucent shading around each line indicates the standard deviation over 5 evaluations on the validation set.}
    \label{fig:naive-vs-corrected-self-training}
\end{figure}
\subsection{Persistent Errors Harm Iterated Self-Training}
\label{sec:persistent-errors}
Applying the formulation of self-training in \cref{sec:rest} results in a improvement, but iterating it further results in program synthesis quality degrading, rather than increasing (\textcolor{red}{red line} in \cref{fig:naive-vs-corrected-self-training}).
This is due to the self-training process inadvertently reinforcing incorrect reasoning.
A program that uses flawed reasoning can occasionally produce a correct answer.
The language model can thus be rewarded for a program that is right for the wrong reasons.
If this goes uncorrected, the language model will learn incorrect reasoning patterns.

We hypothesize that providing a small number of human-written corrections for persistent reasoning errors can stabilize the self-training process.
We use the question type annotations in GQA to identify question types for which training accuracy decreases over time.
These are question types which the language model is not able to self-improve on.
We denote them $\mathcal{Q}_{hard}$.
For each question type in $\mathcal{Q}_{hard}$, we randomly sample one question $q$ for which the language model synthesized a program that produced the wrong answer.
We examine the reasoning in that program, and if the reasoning is flawed, we correct it.
We repeat this until we have a program with correct reasoning for each question type in $\mathcal{Q}_{hard}$, and denote the bank of correct programs as $\mathcal{P}_{gold}$.

We then retrieve from $\mathcal{P}_{gold}$ during self-training for use as in-context examples.
If a question is annotated with a question type in $\mathcal{Q}_{hard}$, we retrieve a correct human-written program from $\mathcal{P}_{gold}$ and use it as an in-context example.
If a question is not annotated with a question type in $\mathcal{Q}_{hard}$, we use a ``default'' in-context example which is the same for all question types not in $\mathcal{Q}_{hard}$.
We show in \cref{fig:naive-vs-corrected-self-training} (\textcolor{ForestGreen}{green line}) that this stabilizes self-training and allows the language model to self-improve across all but a few question types (\cref{fig:question-types-before-after-self-training}).

\subsection{Effect of Data Availability on Self-Training}

\begin{figure}[ht]
    \centering
    \includegraphics[width=\linewidth]{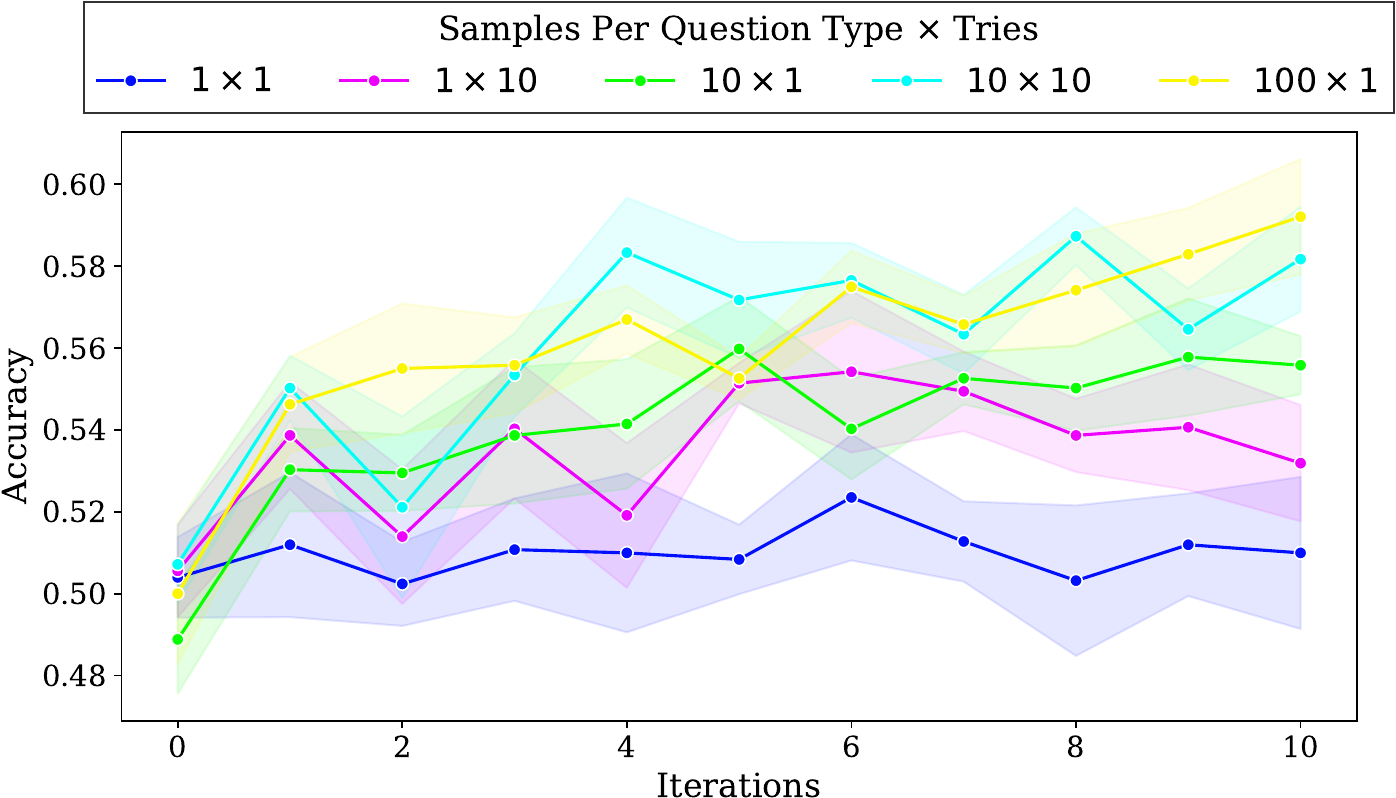}
    \caption{\ourmethod~works even when the amount of available data is reduced by an order of magnitude. We show validation accuracy on GQA. The notation $n \times k$ indicates $n$ samples per question type, with $k$ passes at each sample. For example \textcolor{cyan}{$10 \times 10$} indicates 10 samples per question type, with 10 passes per sample. Although \textcolor{cyan}{$10 \times 10$} has 10x fewer unique samples than \textcolor{Goldenrod}{$100 \times 1$}, there is a $<$ 2\% accuracy difference between them, indicating that more passes per instance can partially mitigate data scarcity.}
    \label{fig:effect-of-data-self-training}
\end{figure}

\textbf{Training With Less Data} We explore this in a controlled setting, by manipulating the number of samples per question type in GQA.
Recall that we originally sample $100$ questions per question type for self-training.
This dataset had $\approx 10$k questions.
We construct a training set with only $10$ and $1$ question per question type, for a total of $\approx 1000$ and $\approx 100$ questions respectively.
Self-training improves upon the baseline (\cref{fig:effect-of-data-self-training}) even when there is an order of magnitude decrease in training data ($100 \rightarrow 10$) .
Only when the amount of available training data is reduced by two orders of magnitude ($100 \rightarrow 1$) does self-training fail to produce an appreciable increase in performance.

\noindent \textbf{Is it possible to mitigate data scarcity?} We previously showed that the benefits of self-training reduce when available data is reduced significantly. 
We now test whether we can mitigate this data scarcity by allowing $\pi_\theta$ multiple attempts at a query $q$ during the \textbf{Grow} step.
Concretely, we allow $\pi_\theta$ a total of $10$ tries at each query under the setting in which we train with 1 and 10 samples per question type, for a total of $1$k and $10$k total samples respectively.
We show in \cref{fig:effect-of-data-self-training} that this mitigates the effect of reduced data. 
Although the data poor \textcolor{magenta}{$1 \times 10$} and \textcolor{cyan}{$10 \times 10$} have $10$x fewer unique questions than \textcolor{green}{$10 \times 1$} and \textcolor{Goldenrod}{$100 \times 1$}, their performance is within a standard deviation of their data rich counterparts.
\subsection{Quantifying Changes in Syntactic Structure}
\begin{figure}[ht]
    \centering
    \includegraphics[width=\linewidth]{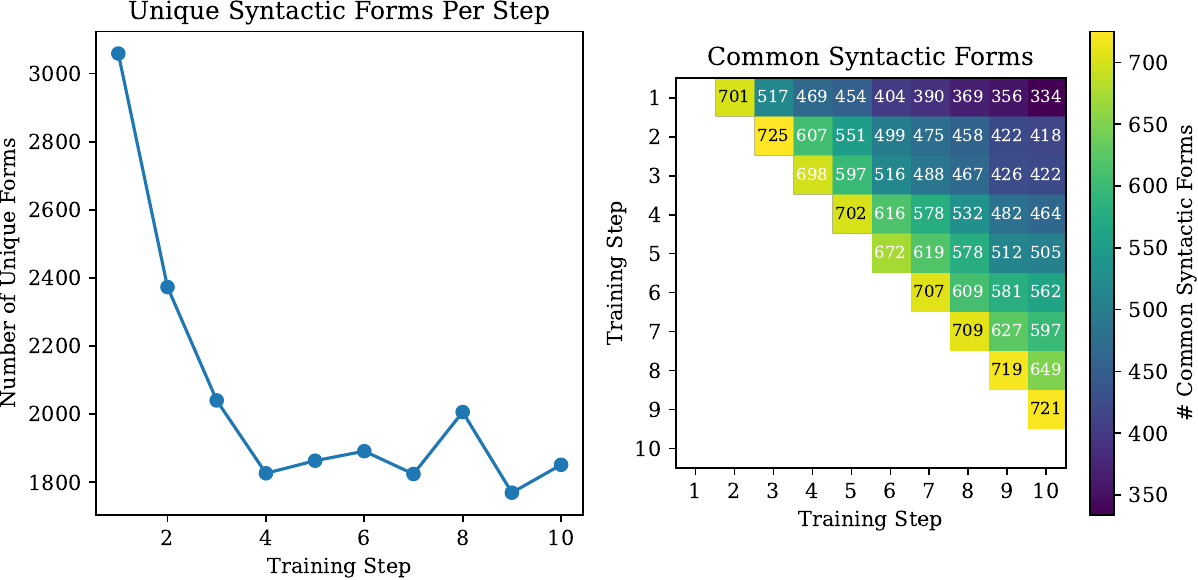}
    \caption{As self-training is iterated, the LLM policy ``hones in'' on a smaller set of syntactic forms, and gradually evolves away from syntactic forms produced by the initial policy. Left Panel: Number of unique normalized abstract syntax trees seen during each iteration of \ourmethod. Right Panel: Number of unique normalized abstract syntax trees in common between each training step. For example, the entry in row 1, column 6 corresponds to the number of unique abstract syntax trees produced by \textit{both} the policy in iteration 1 (initial policy) and the policy in iteration 6.}
    \label{fig:syntactic-analysis}
\end{figure}
How do the programs synthesized by the policy change as self-training is iterated? We examine this by looking at how many unique abstract syntax trees are produced during the \textbf{Grow} step of each iteration.
We parse the synthesized programs into abstract syntax trees, and then normalize the trees to remove irrelevant details such as variable names.
In the left panel of \cref{fig:syntactic-analysis}, we show that the diversity of syntactic forms drops over time.
At the beginning, the policy produces a large number of syntactic forms, but appears to ``hone in'' on a smaller number of forms as self-training continues, and the number of unique syntactic forms drops by almost half.

A remarkably stable set of syntactic forms is conserved from step to step, roughly $\approx 700$ (row above diagonal in right panel of \cref{fig:syntactic-analysis}).
However, the syntactic forms produced by the policy are gradually evolving away from the syntactic forms the initial policy tries, which can be seen in the darkening of the first row in \cref{fig:syntactic-analysis}. 
Despite the coarse reward scheme, the LLM policy gradually explores and learns new syntactic forms.

\begin{table*}[]
\resizebox{\textwidth}{!}{
\begin{tabular}{@{}llcllll@{}}
\toprule
 &     & \multicolumn{1}{c}{VQA} & \multicolumn{2}{c}{Object Detection} & \multicolumn{2}{c}{Image-Text Matching} \\ \midrule
Method & LLM & GQA       & Omnilabel      & Omnilabel-Hard      & Winoground         & SugarCREPE         \\ \midrule
Frozen Proprietary LLM & GPT-3.5-turbo & 53.9 $\pm$ 0.8  & 40.0 $\pm$ 1.2 & 26.0 $\pm$ 1.1 & 45.6 $\pm$ 1.6  & 48.9 $\pm$ 0.7 \\
Frozen Open LLM & CodeLlama-7B  & 50.0 $\pm$ 1.7  & 37.3 $\pm$ 1.6 & 23.7 $\pm$ 1.4 & 41.3 $\pm$ 0.03 &  43.5 $\pm$ 0.8\\
Open LLM + \ourmethod & CodeLlama-7B  & \textbf{59.2 $\pm$ 1.4} & \textbf{42.4 $\pm$ 1.0} & \textbf{28.1 $\pm$ 0.9} & \textbf{52.7 $\pm$ 0.6} &  \textbf{58.7 $\pm$ 1.5} \\ \bottomrule
\end{tabular}
}
\caption{An open LLM policy self-trained with our method substantially outperforms the open policy without self-training, and even outperforms a gpt-3.5-turbo policy. All results use ViperGPT \cite{vipergpt} as the backbone. $\pm$ numbers are the standard deviation over 5 runs. On all datasets except Omnilabel, we report accuracy. On Omnilabel, we report Macro-F1. Higher is better.}
\label{tab:performance}
\end{table*}
\section{Evaluating Functional Correctness}
We measure the functional correctness of the programs synthesized by the self-trained LLM policy $\pi_\theta^\prime$ across three compositional tasks, with the aim of understanding whether:
\begin{enumerate}
    \item Are the programs produced \textit{after} self-training more functionally correct than programs produced \textit{before} self-training?
    \item Is it possible to exceed or match the performance of a much larger proprietary LLM with self-training?
\end{enumerate}

\begin{table}[h!]
\vspace{-3mm}
\resizebox{\linewidth}{!}{
\begin{tabular}{@{}lllll@{}}
\toprule
                                           & \multicolumn{2}{c}{X-Dataset Generalization} & \multicolumn{2}{c}{X-Task Generalization} \\ 
\cmidrule(l{0.5em}r{0.5em}){2-3} \cmidrule(l{0.5em}r{0.5em}){4-5}
Code Generator                  & VQAv2                   & OK-VQA                 & Omnilabel             & SugarCrepe            \\ \midrule
Frozen Few-Shot (7B)      & 46.6 $\pm$ 1.1            & 12.6 $\pm$ 2.1           & 37.3 $\pm$ 1.6          & 43.5 $\pm$ 0.8          \\
\ourmethod~on GQA (7B)  & \textbf{61.0} $\pm$ 1.0            & \textbf{33.7} $\pm$ 1.9           & \textbf{39.9} $\pm$ 0.7          & \textbf{51.0} $\pm$ 1.5          \\
\midrule
\ourmethod~Advantage & +14.4 & +21.1 & +2.6 & +7.5 \\ \bottomrule
\vspace{-5mm}
\end{tabular}
}
\caption{\small{\ourmethod~improves benchmark agnostic visual program synthesis}. A policy self-trained on GQA with \ourmethod~writes better programs for other VQA datasets and other task types. }
\vspace{-3mm}
\label{tab:cross-ds-task-generalization}
\end{table}

For compositional VQA, we use the GQA \cite{gqa} dataset for the reasons outlined in \cref{sec:impl}.
For complex object detection, we choose Omnilabel \cite{omnilabel}.
Omnilabel contains 28K free-form object descriptions over 25K images, and is a challenging task for existing open-vocabulary object detectors due to the complexity of the object descriptions.
For compositional image-text matching, we choose WinoGround \cite{winoground} and SugarCrepe \cite{sugarcrepe}.
State-of-art vision-language models have trouble reaching above chance accuracy on WinoGround, but SugarCrepe is substantially easier.
However, both of these tasks pose significant problems for the \texttt{ImagePatch} API, because many of the relationships mentioned in the text are challenging to detect with the available perception modules.
For all experiments, we use ViperGPT\cite{vipergpt} as the backbone and adopt their prompts.
\textbf{Due to space limitations, many experimental details are in the supplement. }

\subsection{Experimental Setup}
For each task, we apply \ourmethod~as described in \cref{sec:rest}, and evaluate on a held-out subset.
For a comparison with a large proprietary LLM, we use gpt-3.5-turbo.
We evaluate on a subsampled version of each dataset to reduce token costs.
Every LLM is provided the same prompts. 
Each prompt consists of the \texttt{ImagePatch} API specification used in ViperGPT \cite{vipergpt}, and $3$ in-context examples for each task except for object detection, for which we provide $5$ in-context examples.

We use GQA as described in \cref{sec:impl}.
We prepare a compositional subset of Omnilabel \cite{omnilabel} by filtering out all descriptions less than two words in length. 
We then sample a subset of $500$ for evaluation, and a subset of $500$ for training.
To prepare Omnilabel-Hard, we use run a state of the art open-vocabulary object detector (GroundingDINO \cite{groundingdino}) on the remaining OmniLabel samples, and select those which GroundingDINO completely fails on (no detections) to obtain a hard slice.
We then sample a subset of $500$ from the hard slice for evaluation.
For SugarCrepe \cite{sugarcrepe}, we sample $100$ positives and their associated negatives from each of the $6$ categories, for a total of $600$ balanced image-text pairs for validation.
We sample $100$ of the remaining instances from each category for training.
We use all of WinoGround, as it is small enough that there is no need to subsample it.
On WinoGround\cite{winoground}, we evaluate the policy trained on SugarCrepe rather than training on it.
For VQAv2, we sample 10 questions for each of the top-50 most common answers from the compositional subset curated by \cite{vqa_introspect}.

Examples of the inputs for each task are in \cref{fig:before-after}.
We use nucleus sampling with identical parameters for all local LLMs.
We use the API default temperature for gpt-3.5-turbo. 
More details are in the supplement.

\subsection{Discussion}
Across all three tasks, the policy trained by \ourmethod~outperforms both the gpt-3.5-turbo policy, and the initial CodeLlama-7b policy (\cref{tab:performance}).
On GQA, the self-trained policy achieves an absolute improvement of almost $9\%$ over the initial policy, and $5\%$ over the gpt-3.5-turbo policy.
On Omnilabel, self-training produces a $5$\% improvement in Macro-F1 score with only $500$ training samples.
On Omnilabel-Hard, we demonstrate that the visual program synthesis paradigm can localize objects that state of the art open-vocabulary object detectors are unable to localize (Omnilabel-Hard was constructed by selecting instances GroundingDino\cite{groundingdino}) cannot localize).
Even on Omnilabel-Hard, the self-trained policy outperforms the others.
WinoGround and SugarCrepe are difficult to solve by visual program synthesis because many of the relationships are hard to detect with the available perception modules.
Despite the intrinsic difficulty of compositional image-text matching for the \texttt{ImagePatch} API, \ourmethod~produces an increase of $+15$\% over the baseline policy.
The policy trained on SugarCrepe transfers to WinoGround, outperforming the baseline policy by $+10$\%.

\section{Conclusion \& Future Work}
While few-shot prompting of LLMs for visual program synthesis has produced impressive results, it has limitations, because writing good visual programs requires experience with the visual world and the perception modules at ones disposal.
We presented \ourmethod, which improves a LLM's program synthesis abilities using feedback from executing visual programs.
We showed that \ourmethod~produces strong increases over baseline across multiple tasks, and is competitive with gpt-3.5-turbo.
Our work constructed a coarse-valued reward from existing vision-language annotations.
Methods like RLAIF \cite{rlaif}, ReST \cite{rest}, and CodeRL \cite{coderl} all rely on a neural reward model that can provide fine-grained rewards.
Learning from fine-grained rewards is much easier than learning from coarse rewards.
An interesting direction for future work would be to train a neural reward model for visual program synthesis.
Such a reward model could provide fine-grained rewards, and open a broader range of reinforcement learning methods.

{\small
\bibliographystyle{ieeenat_fullname}
\bibliography{11_references}

\begin{thebibliography}{34}
\providecommand{\natexlab}[1]{#1}
\providecommand{\url}[1]{\texttt{#1}}
\expandafter\ifx\csname urlstyle\endcsname\relax
  \providecommand{\doi}[1]{doi: #1}\else
  \providecommand{\doi}{doi: \begingroup \urlstyle{rm}\Url}\fi

\bibitem[Austin et~al.(2021)Austin, Odena, Nye, Bosma, Michalewski, Dohan, Jiang, Cai, Terry, Le, and Sutton]{program_synthesis_with_llms}
Jacob Austin, Augustus Odena, Maxwell Nye, Maarten Bosma, Henryk Michalewski, David Dohan, Ellen Jiang, Carrie~J. Cai, Michael Terry, Quoc~V. Le, and Charles Sutton.
\newblock Program synthesis with large language models.
\newblock \emph{ArXiv}, abs/2108.07732, 2021.

\bibitem[Bai et~al.(2022)Bai, Kadavath, Kundu, Askell, Kernion, Jones, Chen, Goldie, Mirhoseini, McKinnon, Chen, Olsson, Olah, Hernandez, Drain, Ganguli, Li, Tran-Johnson, Perez, Kerr, Mueller, Ladish, Landau, Ndousse, Luko{\vs}iūtė, Lovitt, Sellitto, Elhage, Schiefer, Mercado, DasSarma, Lasenby, Larson, Ringer, Johnston, Kravec, Showk, Fort, Lanham, Telleen-Lawton, Conerly, Henighan, Hume, Bowman, Hatfield-Dodds, Mann, Amodei, Joseph, McCandlish, Brown, and Kaplan]{rlaif}
Yuntao Bai, Saurav Kadavath, Sandipan Kundu, Amanda Askell, John Kernion, Andy Jones, Anna Chen, Anna Goldie, Azalia Mirhoseini, Cameron McKinnon, Carol Chen, Catherine Olsson, Christopher Olah, Danny Hernandez, Dawn Drain, Deep Ganguli, Dustin Li, Eli Tran-Johnson, E Perez, Jamie Kerr, Jared Mueller, Jeff Ladish, J Landau, Kamal Ndousse, Kamilė Luko{\vs}iūtė, Liane Lovitt, Michael Sellitto, Nelson Elhage, Nicholas Schiefer, Noem'i Mercado, Nova DasSarma, Robert Lasenby, Robin Larson, Sam Ringer, Scott Johnston, Shauna Kravec, Sheer~El Showk, Stanislav Fort, Tamera Lanham, Timothy Telleen-Lawton, Tom Conerly, T.~J. Henighan, Tristan Hume, Sam Bowman, Zac Hatfield-Dodds, Benjamin Mann, Dario Amodei, Nicholas Joseph, Sam McCandlish, Tom~B. Brown, and Jared Kaplan.
\newblock Constitutional ai: Harmlessness from ai feedback.
\newblock \emph{ArXiv}, abs/2212.08073, 2022.

\bibitem[Butt et~al.(2024)Butt, Manczak, Wiggers, Rainone, Zhang, Defferrard, and Cohen]{codeit}
Natasha Butt, Blazej Manczak, Auke~J. Wiggers, Corrado Rainone, David~W. Zhang, Michael Defferrard, and Taco Cohen.
\newblock Codeit: Self-improving language models with prioritized hindsight replay.
\newblock \emph{ArXiv}, abs/2402.04858, 2024.

\bibitem[Chen et~al.(2021)Chen, Tworek, Jun, Yuan, de~Oliveira~Pinto, Kaplan, Edwards, Burda, Joseph, Brockman, Ray, Puri, Krueger, Petrov, Khlaaf, Sastry, Mishkin, Chan, Gray, Ryder, Pavlov, Power, Kaiser, Bavarian, Winter, Tillet, Such, Cummings, Plappert, Chantzis, Barnes, Herbert-Voss, Guss, Nichol, Paino, Tezak, Tang, Babuschkin, Balaji, Jain, Saunders, Hesse, Carr, Leike, Achiam, Misra, Morikawa, Radford, Knight, Brundage, Murati, Mayer, Welinder, McGrew, Amodei, McCandlish, Sutskever, and Zaremba]{chen2021codex}
Mark Chen, Jerry Tworek, Heewoo Jun, Qiming Yuan, Henrique~Ponde de Oliveira~Pinto, Jared Kaplan, Harri Edwards, Yuri Burda, Nicholas Joseph, Greg Brockman, Alex Ray, Raul Puri, Gretchen Krueger, Michael Petrov, Heidy Khlaaf, Girish Sastry, Pamela Mishkin, Brooke Chan, Scott Gray, Nick Ryder, Mikhail Pavlov, Alethea Power, Lukasz Kaiser, Mohammad Bavarian, Clemens Winter, Philippe Tillet, Felipe~Petroski Such, Dave Cummings, Matthias Plappert, Fotios Chantzis, Elizabeth Barnes, Ariel Herbert-Voss, William~Hebgen Guss, Alex Nichol, Alex Paino, Nikolas Tezak, Jie Tang, Igor Babuschkin, Suchir Balaji, Shantanu Jain, William Saunders, Christopher Hesse, Andrew~N. Carr, Jan Leike, Josh Achiam, Vedant Misra, Evan Morikawa, Alec Radford, Matthew Knight, Miles Brundage, Mira Murati, Katie Mayer, Peter Welinder, Bob McGrew, Dario Amodei, Sam McCandlish, Ilya Sutskever, and Wojciech Zaremba.
\newblock Evaluating large language models trained on code.
\newblock 2021.

\bibitem[Dettmers et~al.(2023)Dettmers, Pagnoni, Holtzman, and Zettlemoyer]{qlora}
Tim Dettmers, Artidoro Pagnoni, Ari Holtzman, and Luke Zettlemoyer.
\newblock Qlora: Efficient finetuning of quantized llms.
\newblock \emph{ArXiv}, abs/2305.14314, 2023.

\bibitem[Dong et~al.(2023)Dong, Xiong, Goyal, Pan, Diao, Zhang, Shum, and Zhang]{raft}
Hanze Dong, Wei Xiong, Deepanshu Goyal, Rui Pan, Shizhe Diao, Jipeng Zhang, Kashun Shum, and T. Zhang.
\newblock Raft: Reward ranked finetuning for generative foundation model alignment.
\newblock \emph{ArXiv}, abs/2304.06767, 2023.

\bibitem[Gokhale et~al.(2020)Gokhale, Banerjee, Baral, and Yang]{vqa_lol}
Tejas Gokhale, Pratyay Banerjee, Chitta Baral, and Yezhou Yang.
\newblock {VQA-LOL:} visual question answering under the lens of logic.
\newblock In \emph{ECCV}, 2020.

\bibitem[Gulcehre et~al.(2023)Gulcehre, Paine, Srinivasan, Konyushkova, Weerts, Sharma, Siddhant, Ahern, Wang, Gu, Macherey, Doucet, Firat, and de~Freitas]{rest}
Caglar Gulcehre, Tom~Le Paine, Srivatsan Srinivasan, Ksenia Konyushkova, Lotte Weerts, Abhishek Sharma, Aditya Siddhant, Alexa Ahern, Miaosen Wang, Chenjie Gu, Wolfgang Macherey, A. Doucet, Orhan Firat, and Nando de Freitas.
\newblock Reinforced self-training (rest) for language modeling.
\newblock \emph{ArXiv}, abs/2308.08998, 2023.

\bibitem[Gupta and Kembhavi(2022)]{visprog}
Tanmay Gupta and Aniruddha Kembhavi.
\newblock Visual programming: Compositional visual reasoning without training.
\newblock \emph{ArXiv}, abs/2211.11559, 2022.

\bibitem[Haluptzok et~al.(2022)Haluptzok, Bowers, and Kalai]{LangModelsCanKalai2023}
Patrick~M. Haluptzok, Matthew Bowers, and Adam~Tauman Kalai.
\newblock Language models can teach themselves to program better.
\newblock \emph{ArXiv}, abs/2207.14502, 2022.

\bibitem[Hsieh et~al.(2023)Hsieh, Zhang, Ma, Kembhavi, and Krishna]{sugarcrepe}
Cheng-Yu Hsieh, Jieyu Zhang, Zixian Ma, Aniruddha Kembhavi, and Ranjay Krishna.
\newblock Sugarcrepe: Fixing hackable benchmarks for vision-language compositionality.
\newblock In \emph{Thirty-Seventh Conference on Neural Information Processing Systems Datasets and Benchmarks Track}, 2023.

\bibitem[Hu et~al.(2021)Hu, Shen, Wallis, Allen-Zhu, Li, Wang, and Chen]{lora}
J.~Edward Hu, Yelong Shen, Phillip Wallis, Zeyuan Allen-Zhu, Yuanzhi Li, Shean Wang, and Weizhu Chen.
\newblock Lora: Low-rank adaptation of large language models.
\newblock \emph{ArXiv}, abs/2106.09685, 2021.

\bibitem[Hudson and Manning(2019)]{gqa}
Drew~A. Hudson and Christopher~D. Manning.
\newblock {GQA:} {A} new dataset for real-world visual reasoning and compositional question answering.
\newblock In \emph{{IEEE} Conference on Computer Vision and Pattern Recognition, {CVPR} 2019, Long Beach, CA, USA, June 16-20, 2019}, pages 6700--6709. Computer Vision Foundation / {IEEE}, 2019.

\bibitem[Khan et~al.(2023)Khan, BG, Schulter, Yu, Fu, and Chandraker]{seltda}
Zaid Khan, Vijay~Kumar BG, Samuel Schulter, Xiang Yu, Yun Fu, and Manmohan Chandraker.
\newblock Q: How to specialize large vision-language models to data-scarce vqa tasks? a: Self-train on unlabeled images!
\newblock In \emph{Proceedings of the IEEE/CVF Conference on Computer Vision and Pattern Recognition (CVPR)}, 2023.

\bibitem[Lange et~al.(2012)Lange, Gabel, and Riedmiller]{growing_batch_rl}
Sascha Lange, Thomas Gabel, and Martin Riedmiller.
\newblock \emph{Batch Reinforcement Learning}, pages 45--73.
\newblock Springer Berlin Heidelberg, Berlin, Heidelberg, 2012.

\bibitem[Le et~al.(2022)Le, Wang, Gotmare, Savarese, and Hoi]{coderl}
Hung Le, Yue Wang, Akhilesh~Deepak Gotmare, Silvio Savarese, and Steven Hoi.
\newblock Code{RL}: Mastering code generation through pretrained models and deep reinforcement learning.
\newblock In \emph{Advances in Neural Information Processing Systems}, 2022.

\bibitem[Li et~al.(2022)Li, Li, Xiong, and Hoi]{blip}
Junnan Li, Dongxu Li, Caiming Xiong, and Steven Hoi.
\newblock Blip: Bootstrapping language-image pre-training for unified vision-language understanding and generation.
\newblock In \emph{ICML}, 2022.

\bibitem[Liu et~al.(2023)Liu, Zeng, Ren, Li, Zhang, Yang, Li, Yang, Su, Zhu, et~al.]{groundingdino}
Shilong Liu, Zhaoyang Zeng, Tianhe Ren, Feng Li, Hao Zhang, Jie Yang, Chunyuan Li, Jianwei Yang, Hang Su, Jun Zhu, et~al.
\newblock Grounding dino: Marrying dino with grounded pre-training for open-set object detection.
\newblock \emph{arXiv preprint arXiv:2303.05499}, 2023.

\bibitem[Loshchilov and Hutter(2019)]{adamw}
Ilya Loshchilov and Frank Hutter.
\newblock Decoupled weight decay regularization.
\newblock In \emph{International Conference on Learning Representations}, 2019.

\bibitem[Patil et~al.(2023)Patil, Zhang, Wang, and Gonzalez]{gorillallm}
Shishir~G. Patil, Tianjun Zhang, Xin Wang, and Joseph~E. Gonzalez.
\newblock Gorilla: Large language model connected with massive apis.
\newblock \emph{arXiv preprint arXiv:2305.15334}, 2023.

\bibitem[Qin et~al.(2023)Qin, Liang, Ye, Zhu, Yan, Lu, Lin, Cong, Tang, Qian, Zhao, Tian, Xie, Zhou, Gerstein, Li, Liu, and Sun]{toolllm}
Yujia Qin, Shi Liang, Yining Ye, Kunlun Zhu, Lan Yan, Ya-Ting Lu, Yankai Lin, Xin Cong, Xiangru Tang, Bill Qian, Sihan Zhao, Runchu Tian, Ruobing Xie, Jie Zhou, Marc~H. Gerstein, Dahai Li, Zhiyuan Liu, and Maosong Sun.
\newblock Toolllm: Facilitating large language models to master 16000+ real-world apis.
\newblock \emph{ArXiv}, abs/2307.16789, 2023.

\bibitem[Radford et~al.(2021)Radford, Kim, Hallacy, Ramesh, Goh, Agarwal, Sastry, Askell, Mishkin, Clark, Krueger, and Sutskever]{clip}
Alec Radford, Jong~Wook Kim, Chris Hallacy, Aditya Ramesh, Gabriel Goh, Sandhini Agarwal, Girish Sastry, Amanda Askell, Pamela Mishkin, Jack Clark, Gretchen Krueger, and Ilya Sutskever.
\newblock Learning transferable visual models from natural language supervision.
\newblock In \emph{Proceedings of the 38th International Conference on Machine Learning, {ICML} 2021, 18-24 July 2021, Virtual Event}, pages 8748--8763. {PMLR}, 2021.

\bibitem[Rozi{\`e}re et~al.(2023)Rozi{\`e}re, Gehring, Gloeckle, Sootla, Gat, Tan, Adi, Liu, Remez, Rapin, Kozhevnikov, Evtimov, Bitton, Bhatt, Ferrer, Grattafiori, Xiong, D'efossez, Copet, Azhar, Touvron, Martin, Usunier, Scialom, and Synnaeve]{codellama}
Baptiste Rozi{\`e}re, Jonas Gehring, Fabian Gloeckle, Sten Sootla, Itai Gat, Xiaoqing Tan, Yossi Adi, Jingyu Liu, Tal Remez, J{\'e}r{\'e}my Rapin, Artyom Kozhevnikov, I. Evtimov, Joanna Bitton, Manish~P Bhatt, Cristian~Cant{\'o}n Ferrer, Aaron Grattafiori, Wenhan Xiong, Alexandre D'efossez, Jade Copet, Faisal Azhar, Hugo Touvron, Louis Martin, Nicolas Usunier, Thomas Scialom, and Gabriel Synnaeve.
\newblock Code llama: Open foundation models for code.
\newblock \emph{ArXiv}, abs/2308.12950, 2023.

\bibitem[Schick et~al.(2023)Schick, Dwivedi-Yu, Dess{\`i}, Raileanu, Lomeli, Zettlemoyer, Cancedda, and Scialom]{toolformer}
Timo Schick, Jane Dwivedi-Yu, Roberto Dess{\`i}, Roberta Raileanu, Maria Lomeli, Luke Zettlemoyer, Nicola Cancedda, and Thomas Scialom.
\newblock Toolformer: Language models can teach themselves to use tools.
\newblock \emph{ArXiv}, abs/2302.04761, 2023.

\bibitem[Schulter et~al.(2023)Schulter, G, Suh, Dafnis, Zhang, Zhao, and Metaxas]{omnilabel}
Samuel Schulter, Vijay Kumar~B G, Yumin Suh, Konstantinos~M. Dafnis, Zhixing Zhang, Shiyu Zhao, and Dimitris Metaxas.
\newblock Omnilabel: A challenging benchmark for language-based object detection.
\newblock In \emph{ICCV}, 2023.

\bibitem[Selvaraju et~al.(2020)Selvaraju, Tendulkar, Parikh, Horvitz, Ribeiro, Nushi, and Kamar]{vqa_introspect}
Ramprasaath~R. Selvaraju, Purva Tendulkar, Devi Parikh, Eric Horvitz, Marco~Tulio Ribeiro, Besmira Nushi, and Ece Kamar.
\newblock Squinting at vqa models: Introspecting vqa models with sub-questions.
\newblock In \emph{Proceedings of the IEEE/CVF Conference on Computer Vision and Pattern Recognition (CVPR)}, 2020.

\bibitem[Singh et~al.(2023)Singh, Co-Reyes, Agarwal, Anand, Patil, Liu, Harrison, Lee, Xu, Parisi, Kumar, Alemi, Rizkowsky, Nova, Adlam, Bohnet, Sedghi, Mordatch, Simpson, Gur, Snoek, Pennington, Hron, Kenealy, Swersky, Mahajan, Culp, Xiao, Bileschi, Constant, Novak, Liu, Warkentin, Qian, Dyer, Neyshabur, Sohl-Dickstein, and Fiedel]{rest_em}
Avi Singh, John~D. Co-Reyes, Rishabh Agarwal, Ankesh Anand, Piyush Patil, Peter~J. Liu, James Harrison, Jaehoon Lee, Kelvin Xu, Aaron Parisi, Abhishek Kumar, Alex Alemi, Alex Rizkowsky, Azade Nova, Ben Adlam, Bernd Bohnet, Hanie Sedghi, Igor Mordatch, Isabelle Simpson, Izzeddin Gur, Jasper Snoek, Jeffrey Pennington, Jiri Hron, Kathleen Kenealy, Kevin Swersky, Kshiteej Mahajan, Laura Culp, Lechao Xiao, Maxwell~L. Bileschi, Noah Constant, Roman Novak, Rosanne Liu, Tris~Brian Warkentin, Yundi Qian, Ethan Dyer, Behnam Neyshabur, Jascha~Narain Sohl-Dickstein, and Noah Fiedel.
\newblock Beyond human data: Scaling self-training for problem-solving with language models.
\newblock \emph{ArXiv}, abs/2312.06585, 2023.

\bibitem[Subramanian et~al.(2023)Subramanian, Narasimhan, Khangaonkar, Yang, Nagrani, Schmid, Zeng, Darrell, and Klein]{modular_vqa_via_cg}
Sanjay Subramanian, Medhini Narasimhan, Kushal Khangaonkar, Kevin Yang, Arsha Nagrani, Cordelia Schmid, Andy Zeng, Trevor Darrell, and Dan Klein.
\newblock Modular visual question answering via code generation.
\newblock In \emph{Proceedings of the 61st Annual Meeting of the Association for Computational Linguistics (Volume 2: Short Papers)}, pages 747--761, Toronto, Canada, 2023. Association for Computational Linguistics.

\bibitem[Sur\'is et~al.(2023)Sur\'is, Menon, and Vondrick]{vipergpt}
D\'idac Sur\'is, Sachit Menon, and Carl Vondrick.
\newblock Vipergpt: Visual inference via python execution for reasoning.
\newblock \emph{Proceedings of IEEE International Conference on Computer Vision (ICCV)}, 2023.

\bibitem[Thrush et~al.(2022)Thrush, Jiang, Bartolo, Singh, Williams, Kiela, and Ross]{winoground}
Tristan Thrush, Ryan Jiang, Max Bartolo, Amanpreet Singh, Adina Williams, Douwe Kiela, and Candace Ross.
\newblock Winoground: Probing vision and language models for visio-linguistic compositionality.
\newblock In \emph{CVPR}, 2022.

\bibitem[Whitehead et~al.(2022)Whitehead, Petryk, Shakib, Gonzalez, Darrell, Rohrbach, and Rohrbach]{reliable_vqa}
Spencer Whitehead, Suzanne Petryk, Vedaad Shakib, Joseph Gonzalez, Trevor Darrell, Anna Rohrbach, and Marcus Rohrbach.
\newblock Reliable visual question answering: Abstain rather than answer incorrectly.
\newblock In \emph{Proceedings of the European Conference on Computer Vision (ECCV)}, 2022.

\bibitem[Williams(1992)]{reinforce}
Ronald~J. Williams.
\newblock Simple statistical gradient-following algorithms for connectionist reinforcement learning.
\newblock \emph{Machine Learning}, 8\penalty0 (3-4):\penalty0 229--256, 1992.

\bibitem[Xie et~al.(2022)Xie, Raghunathan, Liang, and Ma]{icl_bayesian}
Sang~Michael Xie, Aditi Raghunathan, Percy Liang, and Tengyu Ma.
\newblock An explanation of in-context learning as implicit bayesian inference.
\newblock In \emph{International Conference on Learning Representations}, 2022.

\bibitem[Zeng et~al.(2022)Zeng, Attarian, Ichter, Choromanski, Wong, Welker, Tombari, Purohit, Ryoo, Sindhwani, Lee, Vanhoucke, and Florence]{socratic_models}
Andy Zeng, Maria Attarian, Brian Ichter, Krzysztof Choromanski, Adrian Wong, Stefan Welker, Federico Tombari, Aveek Purohit, Michael Ryoo, Vikas Sindhwani, Johnny Lee, Vincent Vanhoucke, and Pete Florence.
\newblock Socratic models: Composing zero-shot multimodal reasoning with language.
\newblock \emph{arXiv}, 2022.

\end{thebibliography}
}

\ifarxiv \clearpage \appendix \onecolumn
In the appendix, we provide implementation details in \cref{sec:implementation-details}, a failure analysis in \cref{sec:failure-analysis}, more qualitative examples in \cref{sec:qualitative-examples}, and prompts in \cref{sec:prompt}.
\section{Implementation Details}
We use ViperGPT \cite{vipergpt} as our ``backbone''.
We follow their implementation of the \texttt{ImagePatch} API almost exactly. 
We remove some modules and functions that were not necessary for the tasks we explore (e.g. \texttt{llm\_query}) is not necessary for our test datasets.
\label{sec:implementation-details}
\subsection{Grow Step}
During the \textbf{Grow} step, we use nucleus sampling to stochastically sample programs from the language model.
We prompt the language model with the \texttt{ImagePatch} API description in \cref{sec:prompt}. 
In the Huggingface library, this corresponds to the following configuration.
We use a \texttt{top\_p} value of 0.9, which allows the model to consider the most probable tokens that cumulatively make up 90\% of the probability mass. 
We set \texttt{top\_k} was set to 0, disabling the top-k filtering and relying solely on nucleus sampling. 
The \texttt{temperature} parameter was set to 0.7. Temperature effects the randomness of token selection, with values lower than 1 resulting in less random selections. 
We increased the \texttt{max\_new\_tokens} from 180 to 320 to accommodate longer outputs, addressing the issue of premature truncation in programmatic responses. 
Because the \texttt{codellama-7b} model did not include a \texttt{<PAD>} token, we re-use the \texttt{<EOS>} token as the pad token.
\subsection{Improve Step}
During each \textbf{Improve} step, we train the language model using LoRA \cite{lora} for a single epoch.
Following \cite{qlora}, we apply LoRA to all fully-connected layers in CodeLlama. 
In the HuggingFace Transformers library, this corresponds to fc1, fc2, k\_proj, v\_proj, q\_proj, out\_proj in each transformer block.
This corresponds to the MLP blocks and the QKV matrices in the transformer.
We use a LoRA rank of $16$, set $\alpha = 32$, and set the LoRA dropout to $0.05$.
During training, we use a batch size of $4$ and the AdamW \cite{adamw} optimizer.
We use an initial learning rate of $0.0002$ and apply a linear learning rate scheduler with a warmup ratio of $0.1$.

During training, we use the following instruction-following template for language modeling:

\begin{lstlisting}
<s>Write a function using Python and the ImagePatch class (above) that could be executed to provide an answer to the query.

Consider the following guidelines:
- Use base Python (comparison, sorting) for basic logical operations, left/right/up/down, math, etc.

Query: <QUERY GOES HERE>
Program:
<PROGRAM GOES HERE>
<\s>
\end{lstlisting}
Note that the first half of the instruction following template (up to \texttt{Program:}) is identical to the end of the prompt used during the \textbf{Grow} step (\cref{sec:prompt}).
We only apply the language modeling loss to the tokens of the program, rather than the ``instruction''.
\subsection{Evaluation Step}
Hyperparameters and prompts are identical to the \textbf{Grow} step. 
Only the datasets change.
We use the same prompt (\cref{sec:prompt}), the same set of in-context examples, and the same hyperparameters.

\section{Failure Analysis}
\label{sec:failure-analysis}
\subsection{Why does accuracy decrease on some question types?}
In \cref{fig:question-types-before-after-self-training}, we show that self-training allows the language model to improve on \textit{almost} all question types. 
What is happening on question types that the language model does not improve on?
In \cref{tab:qtypes-table}, we list those problematic question types and examples of questions from each of the problematic question types.
Almost all of them tend to have boolean answers or provide a choice between several categories.
To understand why self-training can fail on these questions, consider the scenario of a dataset of entirely boolean questions with possible answers $\{yes,no\}$ where each answer occurs with equal probability.
Now consider a language model policy $\pi_\theta$ that synthesizes programs that result in \textit{yes} half the time, and programs that result in \textit{no} half the time.
In such a case, the policy will receive a non-zero reward approximately $25\%$ of the time, regardless of whether the reasoning in the program was correct or not.
This can reinforce incorrect patterns of reasoning.

\begin{table}[]
\centering
\begin{tabular}{@{}lll@{}}
\toprule
Question Type   & Answer Type        & Example                                               \\ \midrule
stateChoose     & Categorical Choose & How is the water today, still or wavy?                \\
twoDifferent    & Boolean            & Is the vest different in color than the seat?         \\
existAttrNotC   & Boolean            & Is there a truck in the scene that is not green?      \\
diffAnimals     & Boolean            & Are these animals of different types?                 \\
sameGender      & Boolean            & Are both the people of the same gender?               \\
existThatNotC  & Boolean            & Is there a bird in the picture that is not walking?       \\
positionVerifyC & Boolean            & Is the man on the right of the picture?               \\
verifyAttrsC    & Boolean            & Is the towel blue and rectangular?                    \\
existC          & Boolean            & Are there sheep in this picture?                      \\
existMaterialC  & Boolean            & Is there a bottle that is made of glass?              \\
relO            & Open-Ended         & The horse is where?                                   \\
twoCommon       & Boolean            & What do both the shoes and the shorts have in common? \\
existAttrNot   & Boolean            & Is there a fire hydrant in the picture that is not white? \\
exist           & Boolean            & Are there tomatoes?                                   \\
sameAnimals     & Boolean            & Are the animals sheep?                                \\
materialChoose & Categorical Choose & What makes up the lid, plastic or stainless steel?        \\ \bottomrule
\end{tabular}
\caption{Examples of question types from \cref{fig:question-types-before-after-self-training} which suffer reduced accuracy after self-training. Almost all of them are either boolean, or require choosing between several categories. In such cases, self-training can reward incorrect reasoning.}
\label{tab:qtypes-table}
\end{table}

\subsection{Failure Modes}
\subsubsection{Incoherent Reasoning}
\begin{figure}[h]
    \centering
    \includegraphics[width=\linewidth]{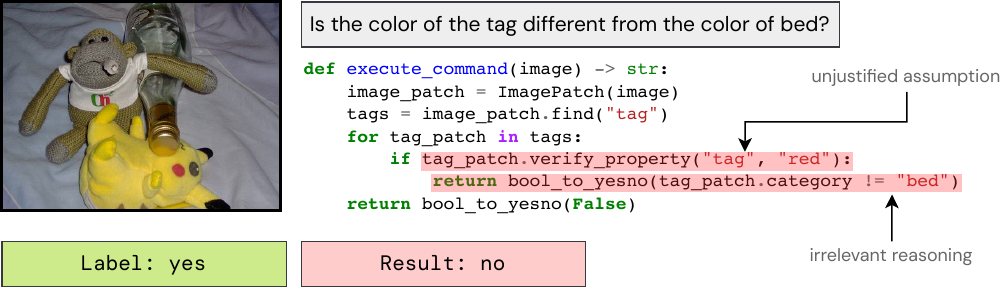}
    \caption{An example of a failure mode in which the LLM employs a line of reasoning which is completely incorrect, using both unjustified assumptions and reasoning irrelevant to the question. This was produced by CodeLlama-7b, but similar errors occur with all LLMs tested.}
    \label{fig:incoherent-reasoning}
\end{figure}

In \cref{fig:incoherent-reasoning}, we show an example of a severe failure mode. 
This failure mode occurs with all evaluated LLMs, including gpt-3.5-turbo.
First, the LLM makes an unjustified assumption, checking to see if the color of the tag is red. 
Second, it compares the \texttt{.category} attribute of the tag to the string ``bed''.
This comparison is irrelevant to the question.
Surprisingly, this failure mode occurs even though the LLM is capable of answering other questions of the same question type which require similar reasoning.
We hypothesize that in situations where the LLM generates completely incoherent reasoning but is able to answer similar questions correctly, further iterations of reinforced self-training will gradually erase this failure mode.
The LLM already ``knows'' how to synthesize the correct program, but needs additional reinforcement.
In situations where the LLM generates completely incoherent reasoning and \textit{is not} able to answer similar questions correctly, we hypothesize that further iterations of reinforced self-training will not erase this failure mode.
One solution in this case is to provide human-written examples of correct reasoning.
As we show in \cref{sec:persistent-errors}, this stabilizes the self-training process.

\subsubsection{Unreliable Perception}
\begin{figure}[h]
    \centering
    \includegraphics[width=\linewidth]{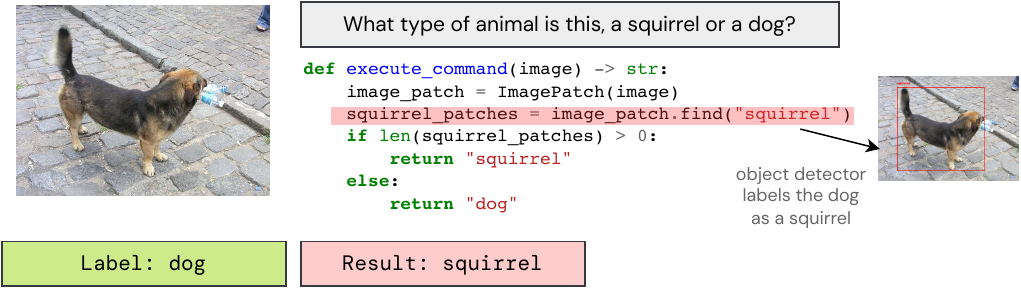}
    \caption{An example of a failure mode in which a perception module is unreliable on a simple input.}
    \label{fig:perception-failure-detection}
\end{figure}

Another type of failure mode is one in which the perception modules are unreliable, as shown in \cref{fig:perception-failure-detection}.
In the case of \cref{fig:perception-failure-detection}, the failure occurs in the \texttt{find} method, which uses GroundingDino as an open vocabulary object detector.
The LLM depends on the \texttt{find} method to return an empty list when ``squirrel'' is not present.
However, the object detector spuriously identifies the dog as a squirrel.

\subsubsection{Complex Relationships}
\begin{figure}[h]
    \centering
    \includegraphics[width=\linewidth]{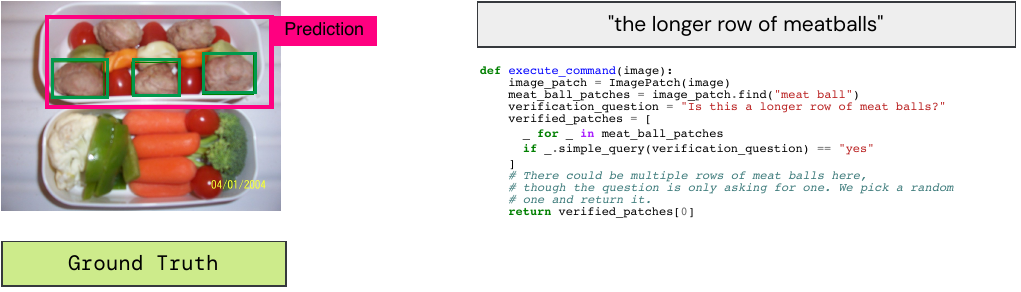}
    \caption{Verifying / detecting complex relationships is challenging for the program synthesis paradigm.}
    \label{fig:complex-rels}
\end{figure}
Another failure mode is one in which the LLM must verify or detect a complex relationship that cannot be handled by the perception modules.
As an example, consider the query in \cref{fig:complex-rels}: ``the longer row of meatballs''.
Recovering the row structure of the meatballs from the detections is not straightforward.
More generally, without a strong visual prior, it is difficult for the LLM to construct a programmatic heuristic for complex relationships.

\section{Qualitative Examples}
In \cref{fig:chatgpt-vs-self-trained,fig:chatgpt-vs-self-trained-v2}, we show examples of visual questions taken from the GQA validation set in which gpt-3.5-turbo (ViperGPT) incorrectly answers queries, but CodeLlama-7B+\ourmethod~does not.
In \cref{fig:ours-vs-gdino-ovod,fig:ours-vs-gdino-ovod-v2}, we show examples in which a state-of-the-art open vocabulary object detector (GroundingDino) is not able to localize described objects, but CodeLlama-7B+\ourmethod~succeeds.
\label{sec:qualitative-examples}
\begin{figure*}
    \centering
    \includegraphics[width=\textwidth]{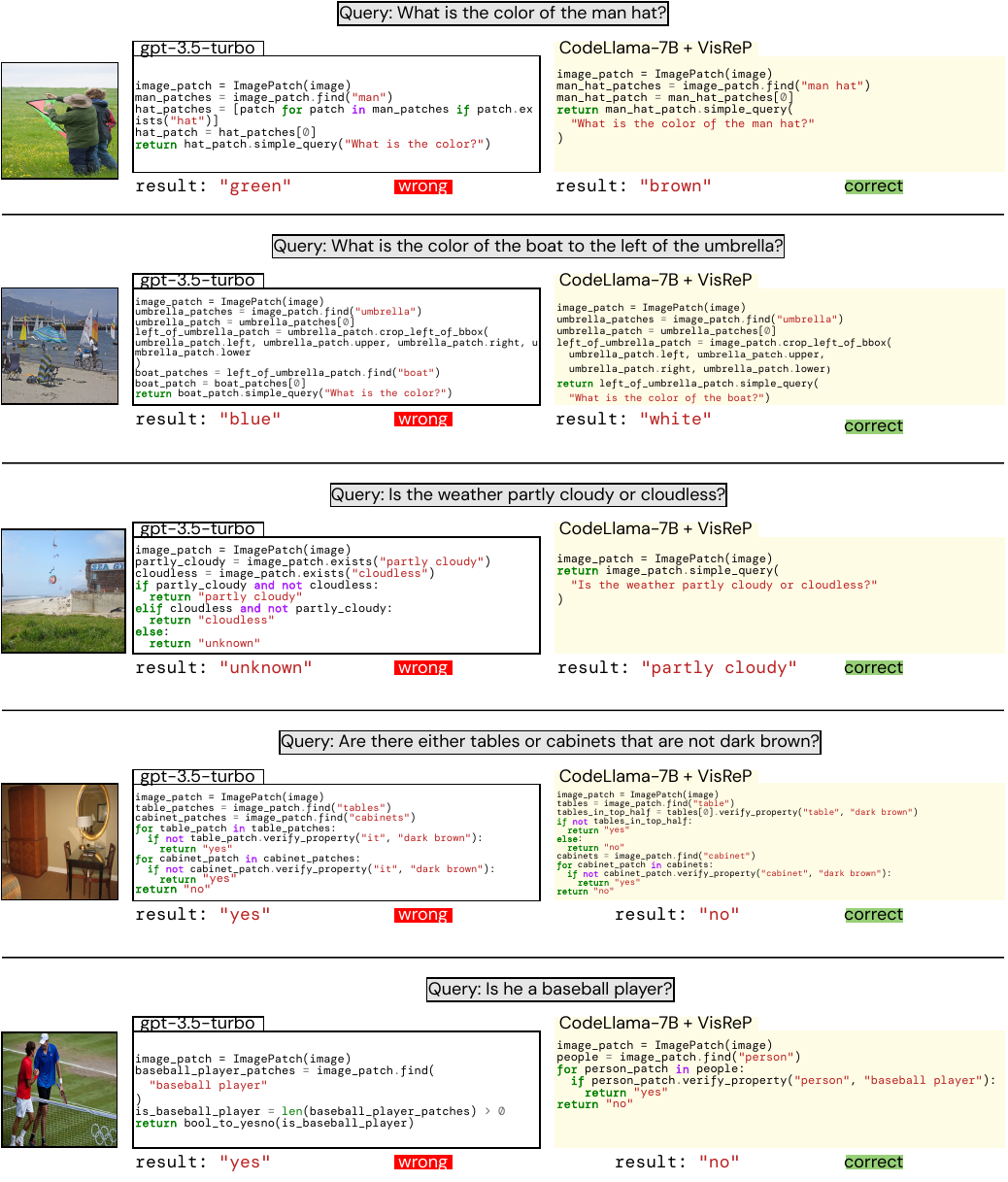}
    \caption{Qualitative examples on VQA (GQA) showing errors made by gpt-3.5-turbo (ViperGPT) that are fixed by \ourmethod.}
    \label{fig:chatgpt-vs-self-trained}
\end{figure*}

\begin{figure*}
    \centering
    \includegraphics[width=\textwidth]{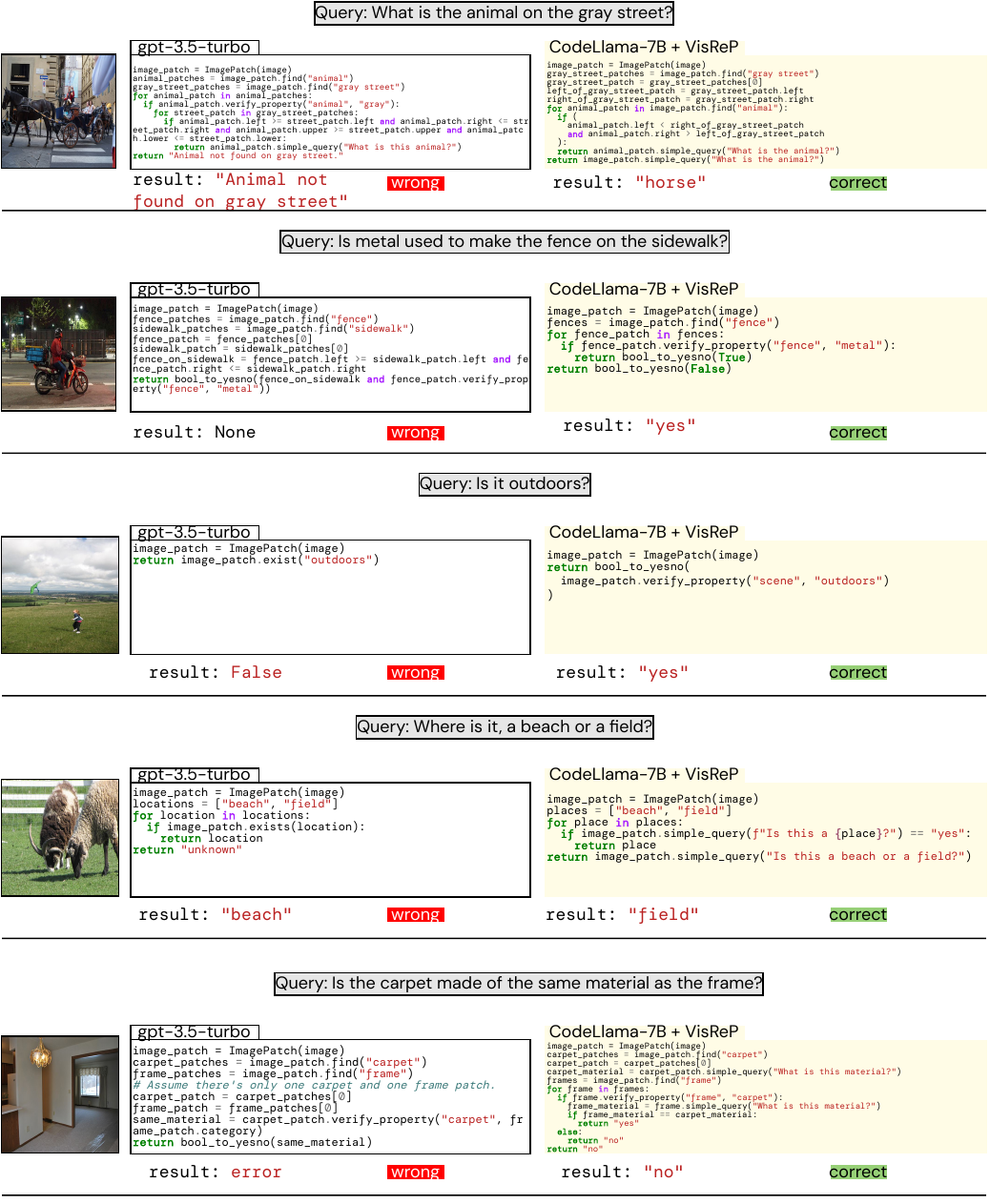}
    \caption{Qualitative examples on VQA (GQA) showing errors made by gpt-3.5-turbo (ViperGPT) that are fixed by \ourmethod.}
    \label{fig:chatgpt-vs-self-trained-v2}
\end{figure*}

\begin{figure*}
    \centering
    \includegraphics[width=0.8\textwidth]{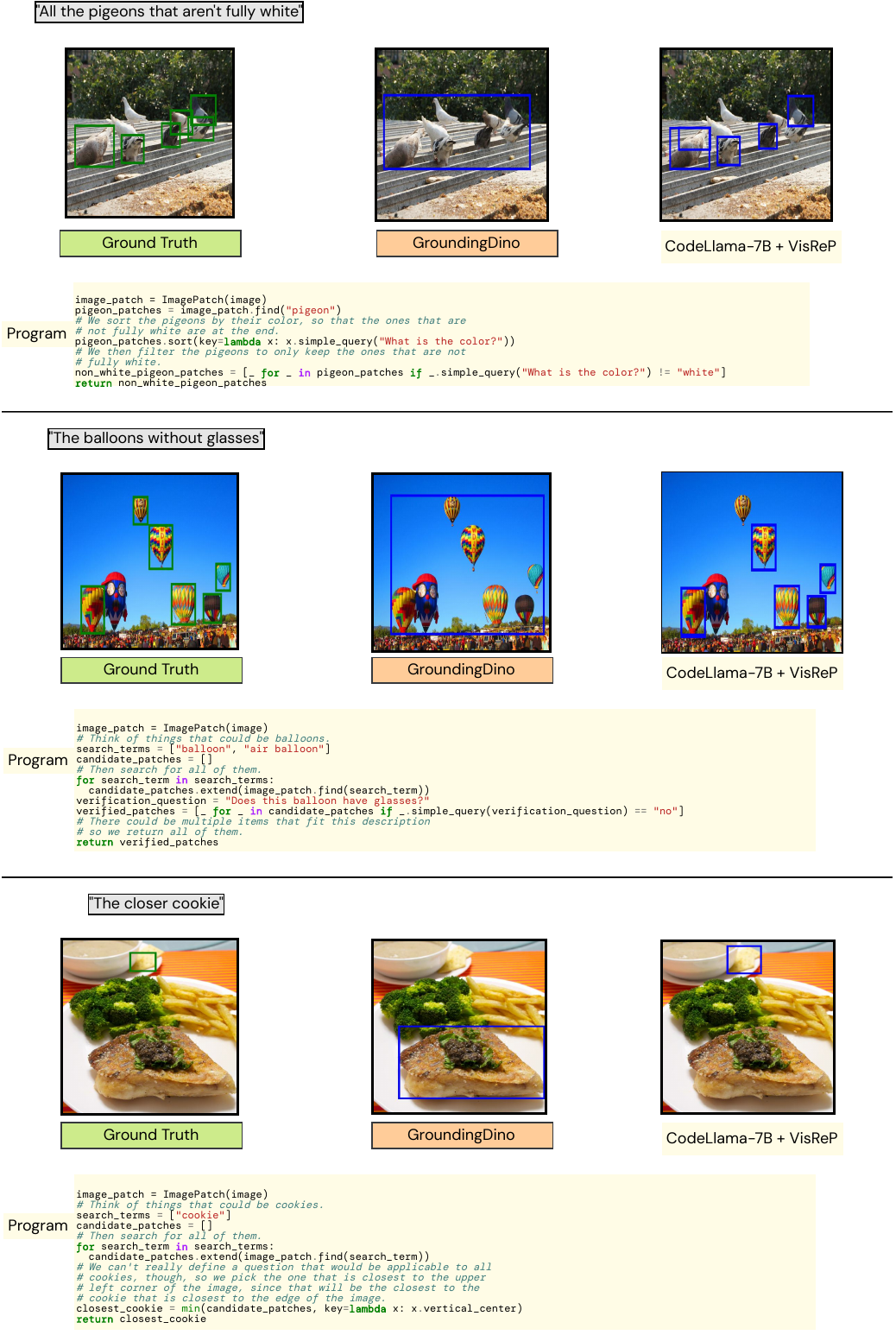}
    \caption{Qualitative examples on object detection (Omnilabel) showing errors made by a state-of-the-art detector (GroundingDino) that are fixed by CodeLlama+\ourmethod.}
    \label{fig:ours-vs-gdino-ovod}
\end{figure*}

\begin{figure*}
    \centering
    \includegraphics[width=0.8\textwidth]{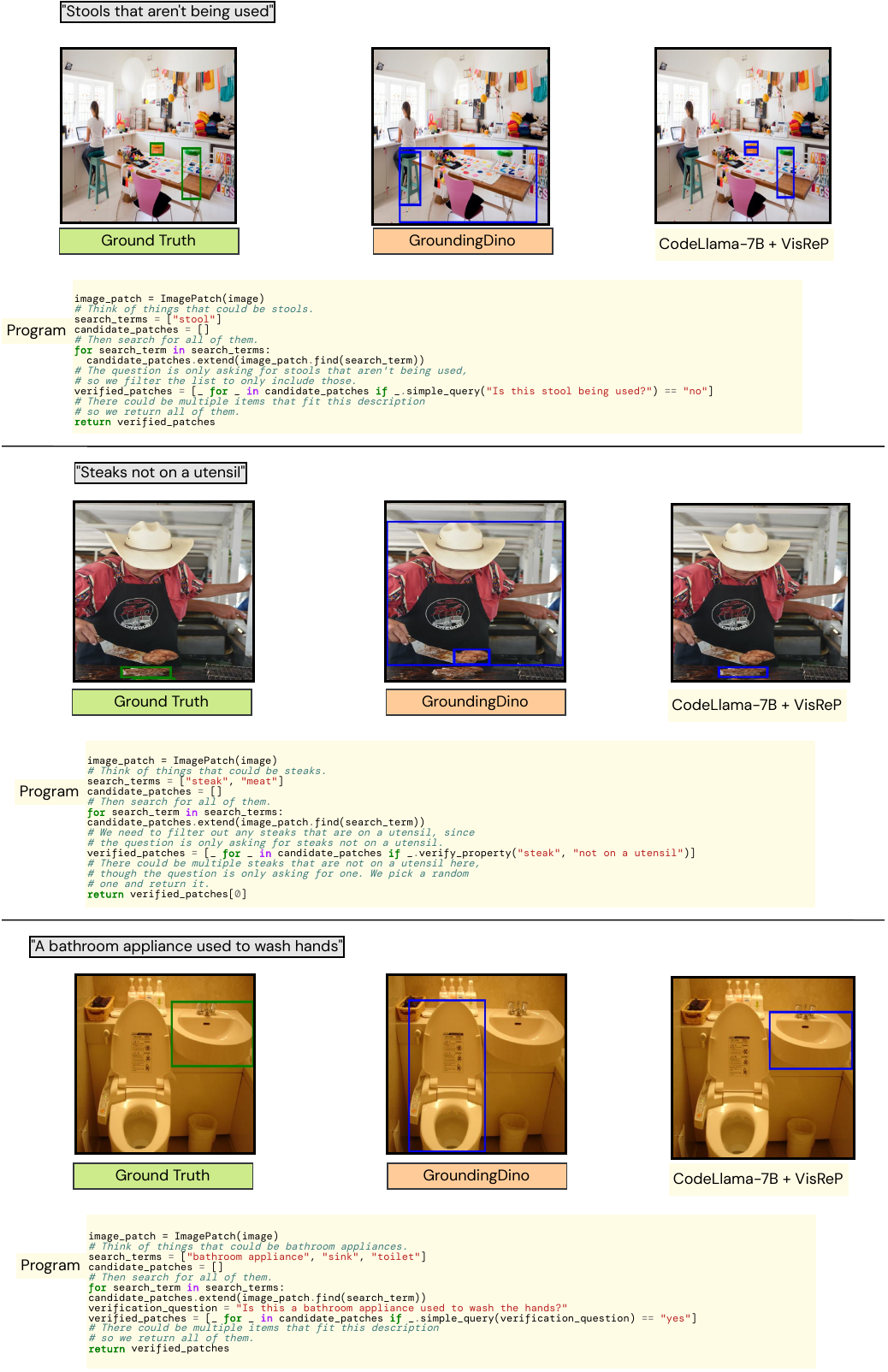}
    \caption{Qualitative examples on object detection (Omnilabel) showing errors made by a state-of-the-art detector (GroundingDino) that are fixed by CodeLlama+\ourmethod.}
    \label{fig:ours-vs-gdino-ovod-v2}
\end{figure*}
\clearpage
\section{ImagePatch API}
\label{sec:prompt}
\begin{lstlisting}[language=Python]
class ImagePatch:
    pass

    def __init__(
        self, image, left=None, lower=None, right=None, upper=None, category=None
    ):
        """Initializes an ImagePatch object by cropping the image at the given
        coordinates and stores the coordinates as attributes. If no coordinates are
        provided, the image is left unmodified, and the coordinates are set to the
        dimensions of the image.
        Parameters
        -------
        image : array_like
            An array-like of the original image.
        left, lower, right, upper : int
            An int describing the position of the (left/lower/right/upper) border of the
             crop's bounding box in the original image.
        category : str
            A string describing the name of the object in the image."""

        self.image = image
        # Rectangles are represented as 4-tuples, (x1, y1, x2, y2),
        # with the upper left corner given first. The coordinate
        # system is assumed to have its origin in the upper left corner, so
        # upper must be less than lower and left must be less than right.

        self.left = left if left is not None else 0
        self.lower = lower if lower is not None else image.height
        self.right = right if right is not None else image.width
        self.upper = upper if upper is not None else 0
        self.cropped_image = image.crop((self.left, self.upper, self.right, self.lower))
        self.horizontal_center = (self.left + self.right) / 2
        self.vertical_center = (self.upper + self.lower) / 2
        self.category = category

    def from_bounding_box(cls, image, bounding_box):
        """Initializes an ImagePatch object by cropping the image at the given
        coordinates and stores the coordinates as attributes.
        Parameters
        -------
        image : array_like
            An array-like of the original image.
        bounding_box : dict
            A dictionary like {"box": [left, lower, right, upper], "category": str}."""
        pass

    @property
    def area(self):
        """
        Returns the area of the bounding box.

        Examples
        --------
        >>> # What color is the largest foo?
        >>> def execute_command(image) -> str:
        >>>     image_patch = ImagePatch(image)
        >>>     foo_patches = image_patch.find("foo")
        >>>     foo_patches.sort(key=lambda x: x.area)
        >>>     largest_foo_patch = foo_patches[-1]
        >>>     return largest_foo_patch.simple_query("What is the color?")
        """
        pass

    def find(self, object_name):
        """Returns a list of ImagePatch objects matching object_name contained in the
        crop if any are found.
        Otherwise, returns an empty list.
        Parameters
        ----------
        object_name : str
            the name of the object to be found

        Returns
        -------
        List[ImagePatch]
            a list of ImagePatch objects matching object_name contained in the crop

        Examples
        --------
        >>> # return the foo
        >>> def execute_command(image) -> List[ImagePatch]:
        >>>     image_patch = ImagePatch(image)
        >>>     foo_patches = image_patch.find("foo")
        >>>     return foo_patches"""
        pass

    def exists(self, object_name):
        """Returns True if the object specified by object_name is found in the image,
        and False otherwise.
        Parameters
        -------
        object_name : str
            A string describing the name of the object to be found in the image.

        Examples
        -------
        >>> # Are there both foos and garply bars in the photo?
        >>> def execute_command(image)->str:
        >>>     image_patch = ImagePatch(image)
        >>>     is_foo = image_patch.exists("foo")
        >>>     is_garply_bar = image_patch.exists("garply bar")
        >>>     return bool_to_yesno(is_foo and is_garply_bar)"""
        pass

    def verify_property(self, object_name, visual_property):
        """Returns True if the object possesses the visual property, and False otherwise.
        Differs from 'exists' in that it presupposes the existence of the object s
        pecified by object_name, instead checking whether the object possesses
        the property.
        Parameters
        -------
        object_name : str
            A string describing the name of the object to be found in the image.
        visual_property : str
            String describing the simple visual property (e.g., color, shape, material)
            to be checked.

        Examples
        -------
        >>> # Do the letters have blue color?
        >>> def execute_command(image) -> str:
        >>>     image_patch = ImagePatch(image)
        >>>     letters_patches = image_patch.find("letters")
        >>>     # Question assumes only one letter patch
        >>>     return bool_to_yesno(letters_patches[0].verify_property("letters", "blue"))
        """
        pass

    def simple_query(self, question):
        """Returns the answer to a basic question asked about the image.
        If no question is provided, returns the answer to "What is this?".
        The questions are about basic perception, and are not meant to be used for
        complex reasoning or external knowledge.
        Parameters
        -------
        question : str
            A string describing the question to be asked.

        Examples
        -------

        >>> # Which kind of baz is not fredding?
        >>> def execute_command(image) -> str:
        >>>     image_patch = ImagePatch(image)
        >>>     baz_patches = image_patch.find("baz")
        >>>     for baz_patch in baz_patches:
        >>>         if not baz_patch.verify_property("baz", "fredding"):
        >>>             return baz_patch.simple_query("What is this baz?")

        >>> # What color is the foo?
        >>> def execute_command(image) -> str:
        >>>     image_patch = ImagePatch(image)
        >>>     foo_patches = image_patch.find("foo")
        >>>     foo_patch = foo_patches[0]
        >>>     return foo_patch.simple_query("What is the color?")

        >>> # Is the second bar from the left quuxy?
        >>> def execute_command(image) -> str:
        >>>     image_patch = ImagePatch(image)
        >>>     bar_patches = image_patch.find("bar")
        >>>     bar_patches.sort(key=lambda x: x.horizontal_center)
        >>>     bar_patch = bar_patches[1]
        >>>     return bar_patch.simple_query("Is the bar quuxy?")"""
        pass

    def visualize(self):
        """Visualizes the bounding box on the original image and annotates it with the category name if provided."""
        pass

    def crop_left_of_bbox(self, left, upper, right, lower):
        """Returns an ImagePatch object representing the area to the left of the given
        bounding box coordinates.

        Parameters
        ----------
        left, upper, right, lower : int
            The coordinates of the bounding box.

        Returns
        -------
        ImagePatch
            An ImagePatch object representing the cropped area.

        Examples
        --------
        >>> # Is the bar to the left of the foo quuxy?
        >>> def execute_command(image) -> str:
        >>>     image_patch = ImagePatch(image)
        >>>     foo_patch = image_patch.find("foo")[0]
        >>>     left_of_foo_patch = image_patch.crop_left_of_bbox(
        >>>         foo_patch.left, foo_patch.upper, foo_patch.right, foo_patch.lower
        >>>     )
        >>>     return bool_to_yesno(left_of_foo_patch.verify_property("bar", "quuxy"))
        """
        pass

    def crop_right_of_bbox(self, left, upper, right, lower):
        """Returns an ImagePatch object representing the area to the right of the given
        bounding box coordinates.

        Parameters
        ----------
        left, upper, right, lower : int
            The coordinates of the bounding box.

        Returns
        -------
        ImagePatch
            An ImagePatch object representing the cropped area.

        Examples
        --------
        >>> # Is the bar to the right of the foo quuxy?
        >>> def execute_command(image) -> str:
        >>>     image_patch = ImagePatch(image)
        >>>     foo_patch = image_patch.find("foo")[0]
        >>>     right_of_foo_patch = image_patch.crop_right_of_bbox(
        >>>         foo_patch.left, foo_patch.upper, foo_patch.right, foo_patch.lower
        >>>     )
        >>>     return bool_to_yesno(right_of_foo_patch.verify_property("bar", "quuxy"))
        """
        pass

    def crop_below_bbox(self, left, upper, right, lower):
        """Returns an ImagePatch object representing the area below the given
        bounding box coordinates.

        Parameters
        ----------
        left, upper, right, lower : int
            The coordinates of the bounding box.

        Returns
        -------
        ImagePatch
            An ImagePatch object representing the cropped area.

        Examples
        --------
        >>> # Is the bar below the foo quuxy?
        >>> def execute_command(image) -> str:
        >>>     image_patch = ImagePatch(image)
        >>>     foo_patch = image_patch.find("foo")[0]
        >>>     below_foo_patch = image_patch.crop_below_bbox(
        >>>         foo_patch.left, foo_patch.upper, foo_patch.right, foo_patch.lower
        >>>     )
        >>>     return bool_to_yesno(below_foo_patch.verify_property("bar", "quuxy"))"""
        pass

    def crop_above_bbox(self, left, upper, right, lower):
        """Returns an ImagePatch object representing the area above the given
        bounding box coordinates.

        Parameters
        ----------
        left, upper, right, lower : int
            The coordinates of the bounding box.

        Returns
        -------
        ImagePatch
            An ImagePatch object representing the cropped area.

        Examples
        --------
        >>> # Is the bar above the foo quuxy?
        >>> def execute_command(image) -> str:
        >>>     image_patch = ImagePatch(image)
        >>>     foo_patch = image_patch.find("foo")[0]
        >>>     above_foo_patch = image_patch.crop_above_bbox(
        >>>         foo_patch.left, foo_patch.upper, foo_patch.right, foo_patch.lower
        >>>     )
        >>>     return bool_to_yesno(above_foo_patch.verify_property("bar", "quuxy"))"""
        pass


def bool_to_yesno(bool_answer: bool) -> str:
    pass


Write a function using Python and the ImagePatch class (above) that could be executed to provide an answer to the query.

Consider the following guidelines:
- Use base Python (comparison, sorting) for basic logical operations, left/right/up/down, math, etc.

INSERT_IN_CONTEXT_EXAMPLES_HERE
Query: INSERT_QUERY_HERE
\end{lstlisting}
 \fi

\end{document}